\documentclass[10pt,twocolumn,letterpaper]{article}

\usepackage{wacv}
\usepackage{times}
\usepackage{epsfig}
\usepackage{graphicx}
\usepackage{amsmath}
\usepackage{amssymb}

\wacvfinalcopy % *** Uncomment this line for the final submission

\ifwacvfinal
\fi

\ifwacvfinal
\usepackage[breaklinks=true,bookmarks=false, hidelinks=true]{hyperref}
\else
\usepackage[pagebackref=true,breaklinks=true,colorlinks,bookmarks=false]{hyperref}
\fi

\usepackage{caption}% http://ctan.org/pkg/caption
\usepackage{glossaries}
\usepackage{comment}

\begin{document}

%%%%%%%%% TITLE
\title{Real-World Super-Resolution of Face-Images from Surveillance Cameras}

\author{
Andreas Aakerberg$^{1}$ \space\space\space Kamal Nasrollahi$^{1,2}$ \space\space\space Thomas B. Moeslund$^{1}$\\
\small $^{1}$Visual Analysis and Perception, Aalborg University, Rendsburggade 14 Aalborg, Denmark\\
\small $^{2}$Research Department, Milestone Systems A/S, Milestone Systems, Brøndby, Denmark\\
{\tt\small \{anaa, tbm\}@create.aau.dk} \space\space\space {\tt\small kna@milestone.dk}

}

\maketitle
\newacronym{cnn}{CNN}{Convolutional Neural Network}
\newacronym{ann}{ANN}{Artificial Neural Network}
\newacronym{sr}{SR}{Super-Resolution}
\newacronym{lr}{LR}{Low-Resolution}
\newacronym{gt}{GT}{Ground-Truth}
\newacronym{hr}{HR}{High-Resolution}
\newacronym{fov}{FOV}{Field-of-View}
\newacronym{ppi}{PPI}{Pixels-per-Inch}
\newacronym{mtf}{MTF}{Modulation Transfer Function}
\newacronym{dof}{DOF}{Depth-of-Field}
\newacronym{snr}{SNR}{Signal-to-Noise Ratio}
\newacronym{ml}{ML}{Maximum Likelihood}
\newacronym{map}{MAP}{Maximum A-Posteriori}
\newacronym{pca}{PCA}{Principal Component Analysis}
\newacronym{ibp}{IBP}{Iterative Back Projection}
\newacronym{nlm}{NL-means}{Non-Local Means}
\newacronym{pmag}{PMAG}{Primary Magnification}
\newacronym{roi}{ROI}{Region of Interest}
\newacronym{psnr}{PSNR}{Peak Signal-to-Noise Ratio}
\newacronym{ssim}{SSIM}{Structural Similarity index}
\newacronym{mse}{MSE}{Mean Squared Error}
\newacronym{mos}{MOS}{Mean Opinion Score}
\newacronym{scn}{SCN}{Sparse Coding Based Network}
\newacronym{caffe}{Caffe}{Convolutional Architecture for Fast Feature Embedding}
\newacronym{sgd}{SGD}{Stochastic Gradient Descent}
\newacronym{relu}{ReLU}{Rectified Linear Units}
\newacronym{sota}{SoTA}{State-of-the-Art}
\newacronym{gpu}{GPU}{Graphics Processing Unit}
\newacronym{tof}{ToF}{Time-of-Flight}
\newacronym{pwm}{PWM}{Pulse-width modulated}
\newacronym{gnn}{GNN}{Global Nearest Neighbor}
\newacronym{fcn}{FCN}{Fully Convolutional Network}
\newacronym{rnn}{RNN}{Recursive Neural Network}
\newacronym{gui}{GUI}{Graphical User Interface}
\newacronym{svm}{SVM}{Support Vector Machine}
\newacronym{ilsvrc}{ILSVRC}{ImageNet Large-Scale Visual Recognition Challenge}
\newacronym{fps}{FPS}{Frames Per Second}
\newacronym{bvlc}{BVLC}{Berkeley Vision and Learning Center}
\newacronym{dsp}{DSP}{Digital Signal Processor}
\newacronym{gmm}{GMM}{General Matrix Multiplication}
\newacronym{cuda}{CUDA}{Compute Unified Device Architecture}
\newacronym{adam}{ADAM}{Adaptive Moment Estimation}
\newacronym{prelu}{PReLU}{Parametric Rectified Linear Unit}
\newacronym{lmdb}{LMDB}{Lightning Memory-Mapped Database}
\newacronym{cad}{CAD}{Computer-Aided Design}
\newacronym{gpc}{GPC}{Gaussian Process Classification}
\newacronym{tsne}{t-SNE}{t-Distributed Stochastic Neighbor Embedding}
\newacronym{gan}{GAN}{Generative Adversarial Network}
\newacronym{mnist}{MNIST}{Mixed National Institute of Standards and Technology}
\newacronym{rgb}{RGB}{Red Green and Blue}
\newacronym{rgbd}{RGB-D}{RGB-Depth}
\newacronym{pcd}{PCD}{Point Cloud Data}
\newacronym{hdd}{HDD}{Hard Disk Drive}
\newacronym{ssd}{SSD}{Solid State Drive}
\newacronym{ir}{IR}{Infrared Radiation}
\newacronym{flops}{FLOPS}{Floating Point Operations per Second}
\newacronym{sift}{SIFT}{Scale Invariant Feature Transform}
\newacronym{surf}{SURF}{Speeded Up Robust Features}
\newacronym{orb}{ORB}{Oriented FAST and Rotated BRIEF}
\newacronym{cudnn}{cuDNN}{CUDA Deep Neural Network Library}
\newacronym{sisr}{SISR}{Single Image Super-Resolution}
\newacronym{misr}{MISR}{Multiple Image Super-Resolution}
\newacronym{mrpa}{MRPA}{Milestone Research Programme at Aalborg University}
\newacronym{ntire}{NTIRE}{New Trends in Image Restoration and Enhancement}
\newacronym{pirm}{PIRM}{Perceptual Image Restoration and Manipulation}
\newacronym{aim}{AIM}{Advances in Image Manipulation}
\newacronym{eccv}{ECCV}{European Conference on Computer Vision}
\newacronym{mor}{MOR}{Mean Opinion Rank}
\newacronym{rwsr}{RWSR}{Real-World Super-Resolution}
\newacronym{pi}{PI}{Perceptual Index}
\newacronym{iqa}{IQA}{Image Quality Assessment}
\newacronym{rrdb}{RRDB}{Residual-in-Residual Dense Block}
\newacronym{vap}{VAP}{Visual Analysis of People}
\newacronym{lfw}{LFW}{Labeled Face in the Wild}
\newacronym{ffhq}{FFHQ}{Flickr-Faces-HQ Dataset}
\newacronym{rafd}{RADF}{Radboud Faces Database}
\newacronym{lpips}{LPIPS}{Learned Perceptual Image Patch Similarity}
\newacronym{nima}{NIMA}{Neural Image Assessment}
\newacronym{dists}{DISTS}{Deep Image Structure and Texture Similarity}
\newacronym{nlpd}{NLPD}{Normalized Laplacian Pyramid Distance}
\newacronym{msssim}{MS-SSIM}{Multi Scale Structural Similarity index}

\begin{abstract}
Most existing face image \gls{sr} methods assume that the \gls{lr} images were artificially downsampled from \gls{hr} images with bicubic interpolation. This operation changes the natural image characteristics and reduces noise. Hence, \gls{sr} methods trained on such data most often fail to produce good results when applied to real \gls{lr} images. To solve this problem, we propose a novel framework for generation of realistic \gls{lr}/\gls{hr} training pairs. Our framework estimates realistic blur kernels, noise distributions, and JPEG compression artifacts to generate \gls{lr} images with similar image characteristics as the ones in the source domain. This allows us to train a \gls{sr} model using high quality face images as \gls{gt}. For better perceptual quality we use a \gls{gan} based \gls{sr} model where we have exchanged the commonly used VGG-loss \cite{vggLoss} with LPIPS-loss \cite{lpips}. Experimental results on both real and artificially corrupted face images show that our method results in more detailed reconstructions with less noise compared to existing \gls{sota} methods. In addition, we show that the traditional non-reference \gls{iqa} methods fail to capture this improvement and demonstrate that the more recent NIMA metric \cite{nima} correlates better with human perception via \gls{mor}. 
\end{abstract}
\glsresetall

\section{Introduction}

\begin{figure}[!h]
\minipage{0.15\textwidth}
        \includegraphics[width=\textwidth]{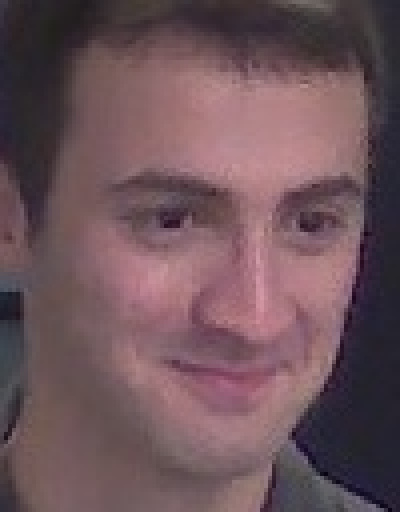}
  \caption*{Original}
\endminipage\hfill
\minipage{0.15\textwidth}
        \includegraphics[width=\textwidth]{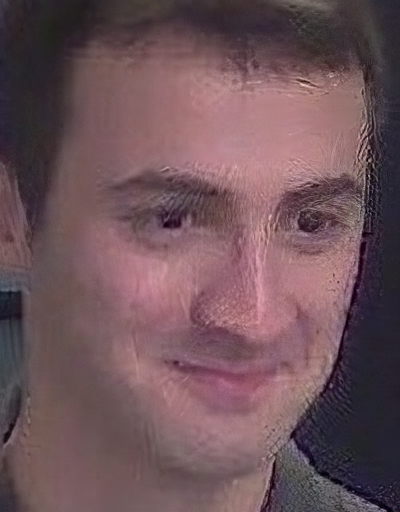}
  \caption*{ESRGAN \cite{esrgan}}
\endminipage\hfill
\minipage{0.15\textwidth}
        \includegraphics[width=\textwidth]{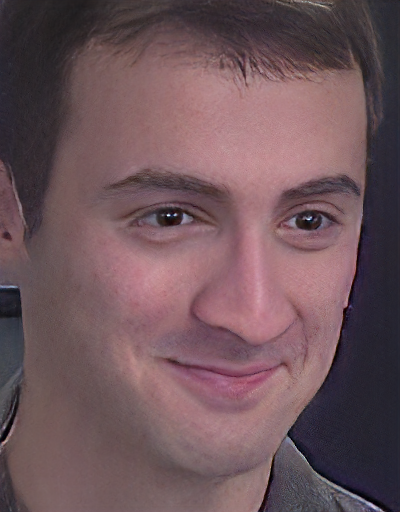}
  \caption*{Ours}
\endminipage
\caption{$\times4$ SR of a real low-quality face image ($100\times128$ pixels) from the Chokepoint DB \cite{chokepoint}. Our method enhances details and removes noise while the ESRGAN \cite{esrgan} amplifies the corruptions.}
\label{fig:goodResult}
\end{figure}

Face \gls{sr} is a special case of \gls{sr} which aims to restore \gls{hr} face images from their \gls{lr} counterparts. This is useful in many different applications such as video surveillance and face enhancement. Current \gls{sota} face \gls{sr} methods based on \glspl{cnn} are able to reconstruct images with photo-realistic appearance from artificially generated \gls{lr} images. However, these methods often assume that the \gls{lr} images were downsampled with bicubic interpolation, and therefore fail to produce good results when applied to real-world \gls{lr} images. This is mostly due to the fact that the downsampling operation with bicubic downscaling changes the natural image characteristics and reduces the amount of artifacts. Hence, when using algorithms trained with supervised learning on such artificial \gls{lr}/\gls{hr} image pairs, the reconstructed images usually contains strong artifacts due to the domain gap. 

This paper is about \gls{sr} of real low-resolution, noisy, and corrupted images, also known as \gls{rwsr}. We apply our proposed method to face images, but the method is also applicable to other image domains. To create a \gls{sr} model that is robust against the corruptions found in real images, we create a degradation framework that can produce \gls{lr} images that have the same image characteristic as the images that we want to super-resolve, \ie the source domain images. By creating \gls{lr} images from clean high-quality images, \ie the target domain, allows us to train a \gls{sr} model that learns to super-resolve images with similar characteristics. This approach is inspired by the work of Ji \etal \cite{realsr} who propose to perform \gls{rwsr} via kernel estimation and noise injection. However, we observe that their framework for image degradation is not ideal for \gls{sr} of \gls{lr} face images from surveillance cameras, as these are often also corrupted by compression artifacts. Hence, we extend the degradation framework from \cite{realsr} to include JPEG compression artifacts. We use the ESRGAN \cite{esrgan} model, which is one of the \gls{sota} models for perceptual quality, as our backbone \gls{sr} model. However, we find that the combination of loss functions for the ESRGAN is not ideal for optimal perceptual quality. To this end, we exchange the VGG-loss \cite{vggLoss} with PatchGAN \cite{cyclegan} loss for the discriminator similar to \cite{realsr}. Inspired by Jo \etal \cite{investigatingLossFunctions}, we additionally exchange the VGG-loss \cite{vggLoss} with \gls{lpips} loss \cite{lpips} for better perceptual quality.
Different from existing models for face \gls{sr} \cite{superFAN, crSR, toLearnSR}, we do not restrict our model to only work for face images of fixed input sizes, which makes our model more useful in practice. To the best of our knowledge, we are the first to propose a method for \gls{sr} of real \gls{lr} face images of arbitrary sizes. 

We evaluate our method on two different datasets. To enable comparison of the \gls{sr} performance against \gls{gt} reference images, we artificially corrupt high-quality face images from \gls{ffhq} \cite{ffhq} and report quantitative results using conventional \gls{iqa} methods and the most recent methods for assessment of the perceptual quality. For evaluation on real \gls{lr} face image from surveillance cameras we use the Chokepoint DB \cite{chokepoint}. In this case, as no \gls{gt} image is available, we report the results using \gls{mor} and several non-reference based \gls{iqa} methods. In both cases we show the effectiveness of our method via quantitative and qualitative evaluations. Furthermore, our evaluations show that most existing non-reference based \gls{iqa} methods correlate poorly with human perception, while the recent \gls{nima} \cite{nima} metric provides a good correlation with human judgment as proven with \gls{mor}.

In summary, our contributions are:
\begin{itemize}
    \item We propose a novel framework for generation of \gls{lr}/\gls{hr} training pairs that includes the most common image degradation types in real-world face images. Our framework includes blur kernel estimation, noise injection and compression artifacts.
    \item We also propose an improved ESRGAN \cite{esrgan} based \gls{sr} model with PatchGAN \cite{cyclegan} and \gls{lpips} loss \cite{lpips} for better perceptual quality, and show the benefit on real \gls{lr} face images from the Chokepoint DB \cite{chokepoint} and artificially corrupted face images from the \gls{ffhq} DB \cite{ffhq}.
    \item Quantitatively, we evaluate our method using the most popular non-reference based \gls{iqa} methods, and find only the recent \gls{nima} \cite{nima} metric to correlate with human judgment via \gls{mor}.
\end{itemize}
\section{Related Work}
Recent advancements within deep-learning have proven very successful for use within super-resolution, and models of this type often achieve \gls{sota} results. The first deep-learning based method for super-resolution was proposed by Dong \etal \cite{srcnn} who successfully trained a \gls{cnn} to learn a non-linear mapping from \gls{lr} to \gls{hr} images. Later proposals relied on deeper networks and residual learning \cite{vdsr, edsr}, recursive learning \cite{drcn}, multi-path learning \cite{dsrn}, and different loss functions \cite{lapsrn} to reduce the reconstruction error between the super-resolved image and the \gls{gt} image. However, while these methods yield high \gls{psnr} values, they tend to produce over-smoothed images which lack high-frequency details. To overcome this, Ledig \etal \cite{srgan} proposed to use \glspl{gan} for \gls{sr} with the SRGAN, to achieve realistic looking images according to human perception. The ESRGAN \cite{esrgan} further improves the SRGAN \cite{srgan} by several changes to the discriminator and generator. 
The \gls{lr} images needed for training the aforementioned deep-learning based super-resolution models are typically created by downsampling \gls{hr} images with an ideal downscaling kernel, typically bicubic downscaling. However, the images generated by this kernel do not nescessarily match real \gls{sr} images. Additionally, in the downscaling process, important natural image characteristics, such as image sensor noise is removed, which the super-resolution algorithms are then prevented from learning. This results in poor reconstruction results and unwanted artifacts when a real-world noisy \gls{lr} image is super-resolved \cite{aim2019}.\\

\noindent \textbf{Real-World Super-Resolution}\space\space\space One way to address the the lack of a proper imaging model for \gls{rwsr}, is to create datasets that consist of real \gls{lr}/\gls{hr} image pairs captured using two cameras with different focal lengths \cite{realsr-dataset, imagepairs, drealsr}. However, this method is cumbersome and has inherent problems with the alignment of the image pairs. To overcome the problem of missing real-world training data, Shocher \etal \cite{zssr} propose a zero-shot approach where a small \gls{cnn} is trained at test time on \gls{lr}/\gls{hr} pairs extracted from the \gls{lr} image itself. Soh \etal \cite{mzsr} extend the work of \cite{zssr} by using meta-transfer learning phase to exploit information from an external dataset. Gu \etal \cite{sftmd} train a kernel estimator and corrector \glspl{cnn} under the assumption that the downscaling kernel belongs to a certain family of Gaussian filters and uses the estimated kernel as input to a super-resolution model. To super-resolve \gls{lr} images with arbitrary blur kernels, Zhang \etal \cite{deep-plug-and-play-sr} propose a deep plug-and-play framework which takes advantage of existing blind deblurring methods for blur kernel estimation. Bell-Kligler \etal \cite{kernelgan} trains a \gls{gan} to estimate blur kernels from \gls{lr} images and combines it with the ZSSR \gls{sr} model \cite{zssr}. Fritsche \etal \cite{dsgan} train a \gls{gan} to introduce natural image characteristics to images downsampled with bicubic downscaling, which is then used to train a super-resolution for improved performance on real-world images. Zhang \etal \cite{usrnet} propose an iterative network for \gls{sr} of blurry, noisy images for different scaling factors by leveraging both learning and model-based methods. Most recently Ji \etal \cite{realsr} propose a degradation framework for the creation of \gls{lr}\gls{hr} image pairs for training. The degradation framework estimates blur kernels and noise distributions from real \gls{lr} images in the source domain which are used to degrade \gls{hr} images in the target domain. This enables training of a \gls{gan} based \gls{sr} model which is shown to perform better on real \gls{lr} images. However, a key limitation of this method is that it does not address the compression artifacts often found in real-world images.\\

\begin{figure}[!h]
\minipage{0.15\textwidth}
        \includegraphics[width=\textwidth]{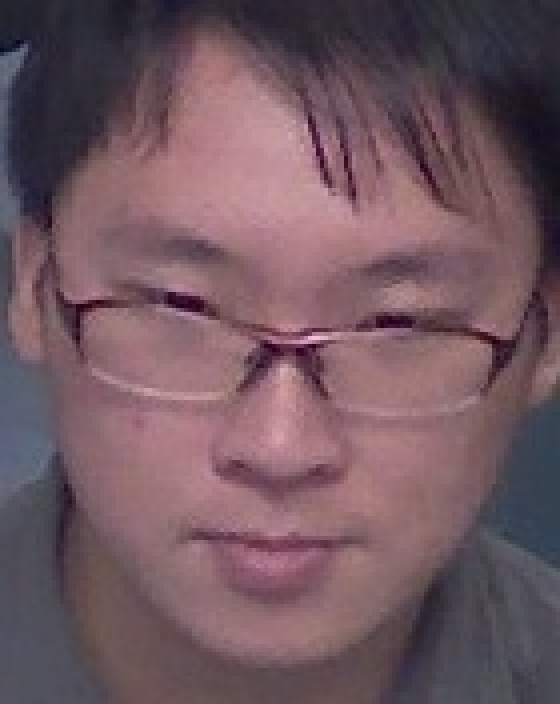}
  \caption*{Original}
\endminipage\hfill
\minipage{0.15\textwidth}
        \includegraphics[width=\textwidth]{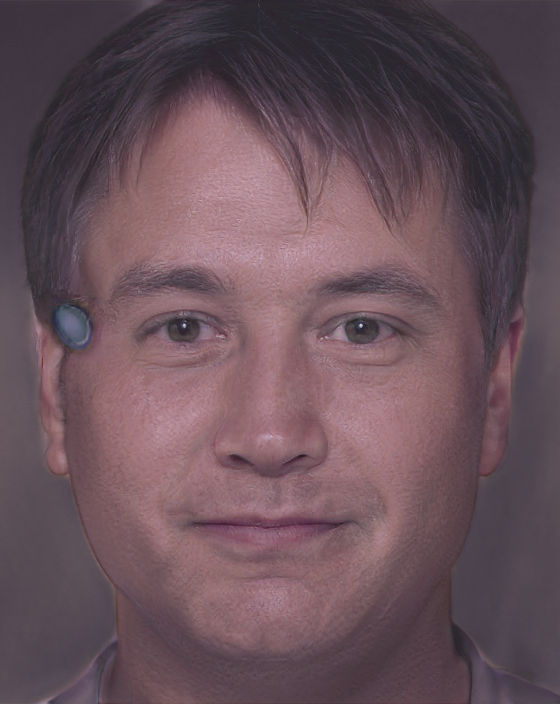}
  \caption*{PULSE \cite{pulse}}
\endminipage\hfill
\minipage{0.15\textwidth}
        \includegraphics[width=\textwidth]{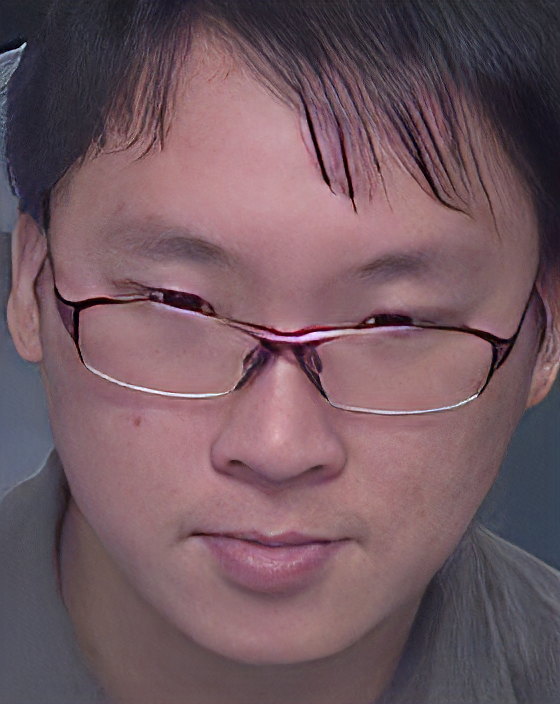}
  \caption*{Ours}
\endminipage
\caption{An example of \gls{sr} of a real low-quality face image from the Chokepoint DB \cite{chokepoint}, where it can be seen that the PULSE \cite{pulse} method changes the identity of the person, while our method preserves the identity and enhances details.}
\label{fig:pulseResult}
\end{figure}

\noindent \textbf{Face Super-Resolution}\space\space\space Face \gls{sr} is a \gls{sr} technique specialized for reconstruction of face images. One of the first methods for face \gls{sr} was proposed by Baker and Kanade \cite{firstFaceSR}. This method reconstructed face details by searching for the most optimal mapping between \gls{lr} and \gls{hr} patches. More recent work relies on deep learning based methods with \glspl{cnn} and \glspl{gan}. Dahl \etal \cite{pixelRecursiveSR} use pixel recursive learning with two \glspl{cnn} to synthesize realistic hair and skin details. Chen \etal \cite{fsrnet} combine face \gls{sr} and face alignment to achieve previously unseen \gls{psnr} values. By searching the latent space of a generative model for images that downscale correctly, Menon \etal \cite{pulse} are able to create face images of high resolution and perceptual quality. However, the problem with this approach is that the generated faces are often far from the true identity of the actual person, as illustrated in Figure~\ref{fig:pulseResult}. Additionally, none of the above mentioned methods are robust against noise or other corruptions in the input images \cite{faceHallucinationRevisited}.     

There are very few publications available in the literature which address the problem of \gls{rwsr} of face-images \cite{faceHallucinationRevisited}. Furthermore, the few existing face \gls{rwsr} methods are only compatible with \gls{lr} images that have been squared to $16\times16$ pixels, meaning that the reconstructed image will be only $64\times64$ or $128\times128$ pixels depending on the scaling factor \cite{superFAN, crSR, toLearnSR}. Hence, these models cannot perform true \gls{sr} directly on the \gls{lr} images. This means that the actual usefulness of the existing face \gls{sr} models is limited. On the contrary, our work presents one possible solution for $\times4$ \gls{rwsr} of face images of arbitrary sizes, which we evaluate on real \gls{lr} face images from surveillance cameras without any prior re-scaling. 

\section{The Proposed Framework}
\begin{figure*}[!h]
\centering
\minipage{0.124\textwidth}
    \centering\captionsetup{margin=0.1cm}
        \includegraphics[width=\textwidth]{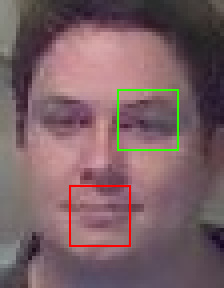}
  \caption*{Original}
\endminipage
\hspace{0.3cm}
\minipage{0.1\textwidth}
    \centering\captionsetup{margin=0.0cm}
        \includegraphics[width=1.4cm]{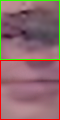}
        \caption*{MZSR\cite{mzsr}}
\endminipage
\hspace{0.3cm}
\minipage{0.1\textwidth}
    \centering\captionsetup{margin=0.0cm}
        \includegraphics[width=1.4cm]{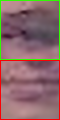}
        \caption*{EDSR\cite{edsr}}
\endminipage
\hspace{0.3cm}
\minipage{0.1\textwidth}
    \centering\captionsetup{margin=0.0cm}
        \includegraphics[width=1.4cm]{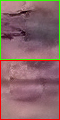}
        \caption*{ESRGAN\cite{esrgan}}
\endminipage
\hspace{0.3cm}
\minipage{0.1\textwidth}
    \centering\captionsetup{margin=0.0cm}
        \includegraphics[width=1.4cm]{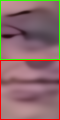}
        \caption*{USRNet\cite{usrnet}}
\endminipage
\hspace{0.3cm}
\minipage{0.1\textwidth}
    \centering\captionsetup{margin=0.0cm}
        \includegraphics[width=1.4cm]{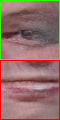}
        \caption*{RealSR\cite{realsr}}
\endminipage
\hspace{0.3cm}
\minipage{0.1\textwidth}
    \centering\captionsetup{margin=0.0cm}
        \includegraphics[width=1.4cm]{figs/closeup-combined-dpsr.png}
        \caption*{DPSR\cite{dpsr}}
\endminipage
\hspace{0.3cm}
\minipage{0.1\textwidth}
    \centering\captionsetup{margin=0.0cm}
        \includegraphics[width=1.4cm]{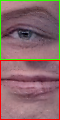}
        \caption*{Ours}
\endminipage
\caption{Comparison with \gls{sota} methods for \gls{sr} of a small face image ($56\times72\:pixels$) from the Chokepoint DB \cite{chokepoint}. As visible, our method hallucinates more realistic face details than the existing methods.}
\label{fig:closeup}
\end{figure*}

This section describes our two-step framework for \gls{rwsr}. The first step aims to generate \gls{lr} images from clean \gls{hr} images in the target domain $Y$, such that these have similar image characteristics as the ones in the source domain $X$. The second step involves training a \gls{sr} model on the constructed paired data, and optimizing for perceptual quality. 

\subsection{Novel Image Degradation}\label{sec:framework}
Traditional approaches for \gls{sr} assumes that a \gls{lr} image $I_{LR}$ is the result of a downscaling operation of the corresponding \gls{hr} image $I_{HR}$ using some kernel $k$ and scaling factor $s$, namely:
\begin{equation}
    I_{LR} = (I_{HR} * k)\downarrow_{s}
\end{equation}
\label{eq:basicSR}
However, real \gls{lr} images from cameras are influenced by multiple other factors that degrade the image as well. The RealSR \cite{realsr} framework tries to address this issue by considering realistic noise distributions and blur kernels in the downscaling process. However, we observe that real images from surveillance cameras are often also degraded with compression artifacts, which makes the RealSR framework perform poorly on such images. To this end, we extend the degradation framework from \cite{realsr} to include JPEG compression artifacts in addition to estimation of realistic noise distributions and blur kernels. Thus, we extend the basic \gls{sr} formulation from Equation~\ref{eq:basicSR}, and assume that the following image degradation model was used to create $I_{LR}$.
\begin{equation}
I_{LR} = (I_{HR} * k)\downarrow_{s} + n + c    
\end{equation}
where $k$, $s$, $n$, and $c$ denotes the blur kernel, scaling factor, noise, and compression artifacts, respectively. $I_{HR}$ is unknown together with $k$, $n$, and $c$. In our degradation framework, we estimate the kernel and noise directly from the images in the source domain $X$. We build a pool of the estimated kernels and noise patches which is used to generate corrupted \gls{lr} images from clean \gls{hr} images and finally JPEG compress the images, in order to create image pairs for training the \gls{sr} model.

\subsection{Blur Kernel Estimation} For estimation of realistic blur kernels, we adopt the KernelGAN method by Bell-Kligler \etal \cite{kernelgan}. This method estimates an image specific \gls{sr} kernel $k_i$ using an unsupervised approach. More specifically, a \gls{gan} is trained to down-scale the input image in a way that best preserves the image patch distributions across scales. We estimate realistic blur kernels from training images in $X$ to form a pool of kernels that can be used to degrade the \gls{hr} images in $Y$.\\

\noindent \textbf{Downsampling}\space\space\space To create the downsampled image $I_D$ we randomly choose a blur kernels $k_i$ from the pool of estimated kernels and perform cross-correlation with images in $Y$. More formally the process is described as:
\begin{equation}
    I_D = (Y_n \ast k_i)\downarrow_s, i\in \{1,2\cdot \cdot \cdot m\}
\end{equation}
where $I_D$ is the downscaled image, $Y_n$ is a \gls{hr} image, $k_i$ refers to a kernel from the degradation pool $\{k_1 , k_2 , \cdot \cdot \cdot k_m\}$ and $s$ is the scaling factor.
\subsection{Noise Estimation}
For degradation with realistic image noise, we adopt the method from \cite{imageBlindDenoising} to extract noise patches from the source images $X$. Here the assumption is that an approximate noise patch can be obtained from a noisy image by extracting an area with weak background and then subtracting the mean. We define two patches $p_i$ and $q_{j}^{i}$. We obtain $p_i$ by a sliding window approach across images in $X$, and similarly for $q_{j}^{i}$ by scanning $p_i$. $p_i$ is considered a smooth patch if the following constraints are met:
\begin{equation}
    |Mean(q_{j}^{i}) - Mean(p_i)| \leq \mu \cdot Mean(p_i)
\end{equation}
and
\begin{equation}
    |Var(q_{j}^{i}) - Var(p_i)| \leq \gamma \cdot Var(p_i)
\end{equation}
where $Mean$ and $Var$ denotes the mean and variance respectively, and $\mu$ and $\gamma$ are scaling factors. Different from \cite{imageBlindDenoising} we add an additional constraint to ensure that saturated patches are not extracted:
\begin{equation}
     Var(p_i) \geq \phi
\end{equation}
where $\phi$ denotes a minimum variance threshold. If all constraints are satisfied, $p_i$ will be considered  a smooth patch. We then create a pool of noise patches $n_i$ by subtracting the mean value from all valid $p_i$.\\ 

\noindent \textbf{Degradation with Noise}\space\space\space We degrade the \gls{lr} images by injecting real noise patches from the noise pool. For better regularization of the \gls{sr} model we randomly pick a noise patch from the noise pool and inject it to the \gls{lr} image during training. The downscaled and noisy \gls{lr} image $I_{N}$ is created as follows:
\begin{equation}
    I_{N} = I_D + n_i, i\in \{1,2\cdot \cdot \cdot l\}
\end{equation}
where $I_D$ is a downscaled image, and $n_i$ is a noise patch from the noise pool $\{n_1 , n_2 , \cdot \cdot \cdot n_l\}$

\subsection{Degradation with Compression artifacts}
Finally, we introduce compression artifacts to the \gls{lr} training images to close the domain gap between these and the real JPEG compressed \gls{lr} images in the source domain $X$. As there are no way of determining the compression strength of existing JPEG images we empirically compare images from $X$ to similar images with different JPEG compression strengths applied and find that a compression strength of 30 results in similar compression artifacts.

\subsection{Backbone Model}
We base our \gls{sr} model on the ESRGAN \cite{esrgan}, which is one of the \gls{sota} networks for perceptual \gls{sr} with $\times4$ upscaling, and train it on the paired \gls{lr} and \gls{hr} images generated with our degradation framework. Different from the SRGAN \cite{srgan}, the ESRGAN uses \glspl{rrdb} in the generator network and the discriminator predicts the relative realness instead of an absolute value. Additionally, the ESRGAN removes the batch normalization layers used in SRGAN.\\ 

\begin{figure*}[!h]
\centering
\minipage{0.1\textwidth}
    \centering\captionsetup{margin=0.0cm}
        \includegraphics[width=\textwidth]{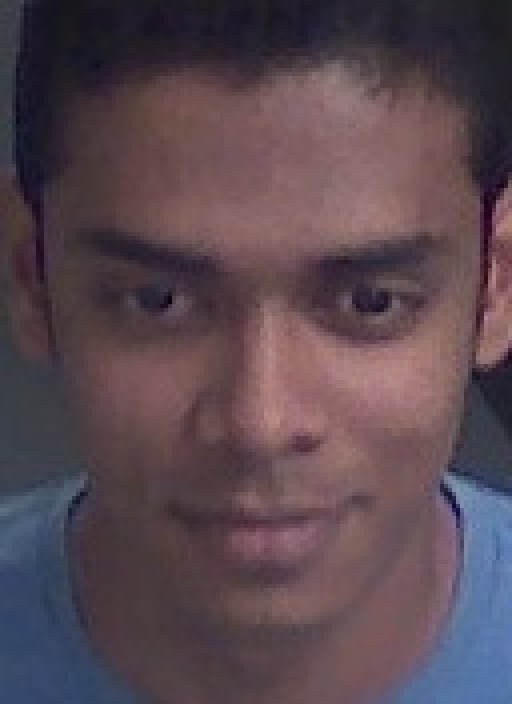}
        \includegraphics[width=\textwidth]{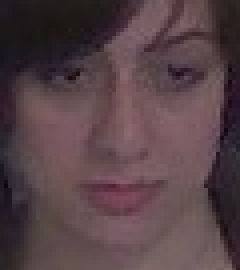}
        \includegraphics[width=\textwidth]{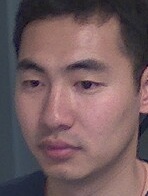}
  \caption*{Original}
\endminipage\hspace{0.3cm}
\minipage{0.1\textwidth}
    \centering\captionsetup{margin=0.0cm}
        \includegraphics[width=\textwidth]{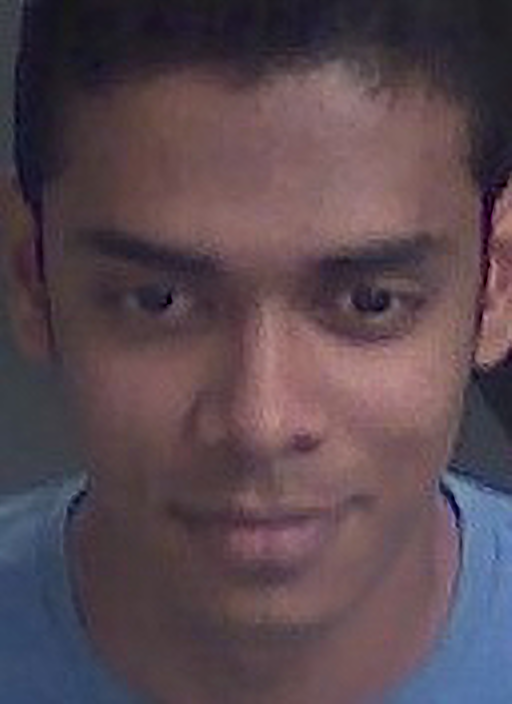}
        \includegraphics[width=\textwidth]{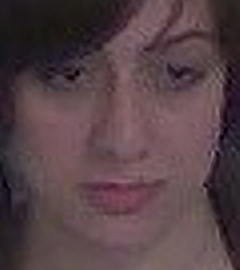}
        \includegraphics[width=\textwidth]{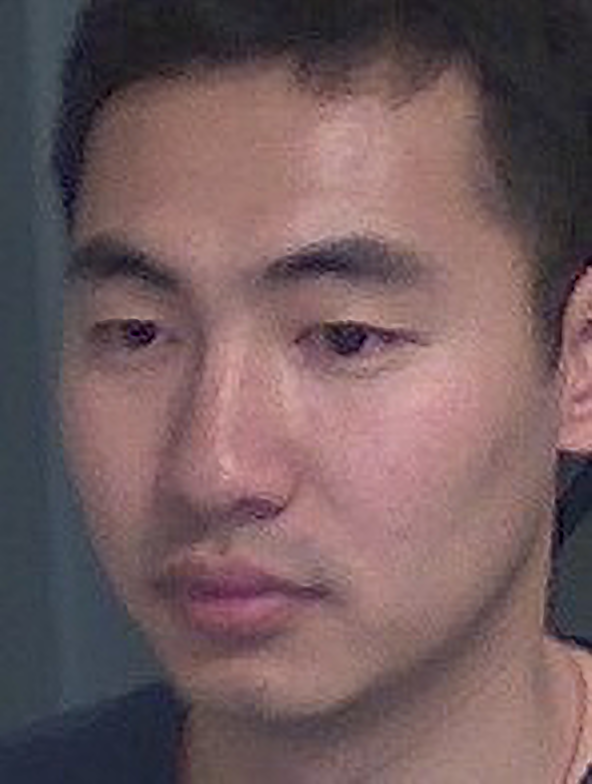}
  \caption*{MZSR\cite{mzsr}}
\endminipage\hspace{0.3cm}
\minipage{0.1\textwidth}
    \centering\captionsetup{margin=0.0cm}
        \includegraphics[width=\textwidth]{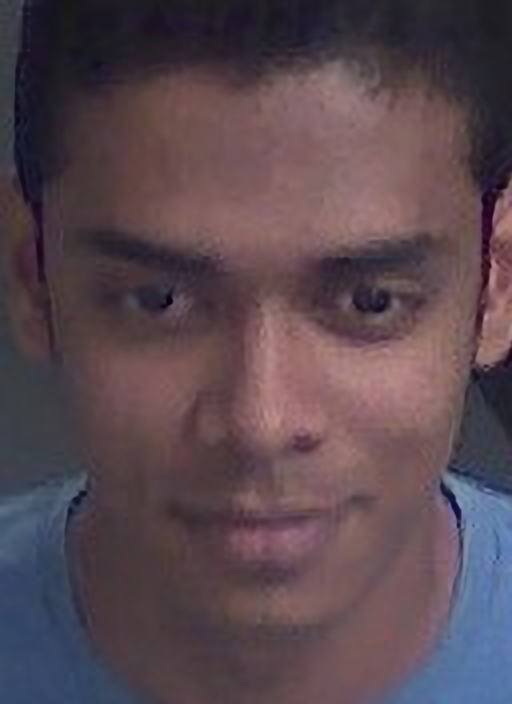}
        \includegraphics[width=\textwidth]{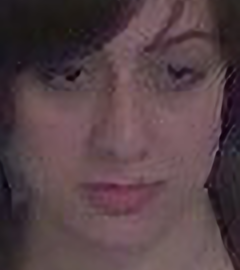}
        \includegraphics[width=\textwidth]{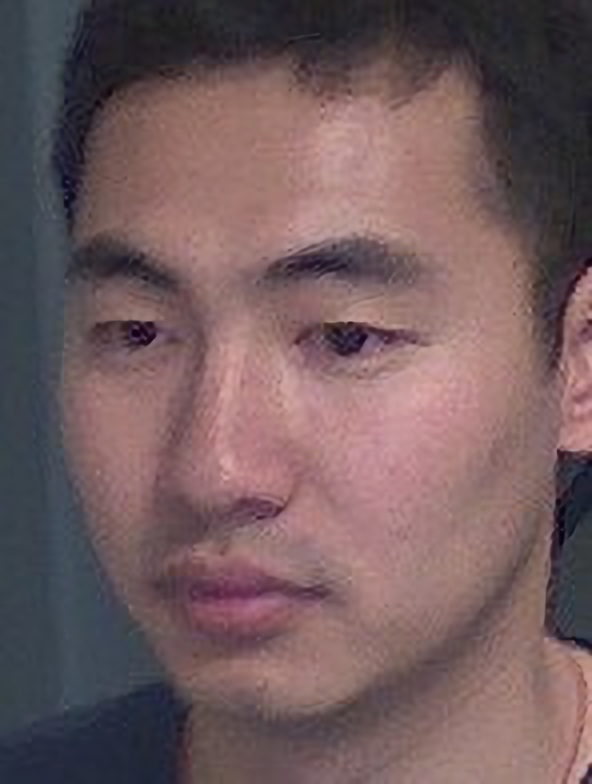}
  \caption*{EDSR\cite{edsr}}
\endminipage\hspace{0.3cm}
\minipage{0.1\textwidth}
    \centering\captionsetup{margin=0.0cm}
        \includegraphics[width=\textwidth]{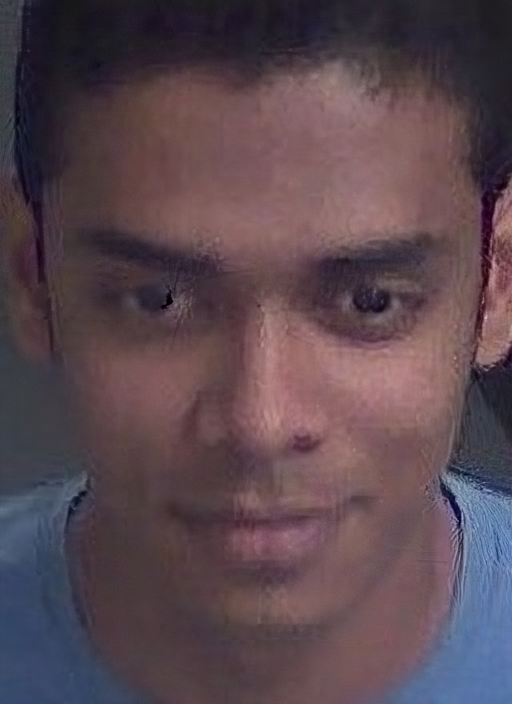}
        \includegraphics[width=\textwidth]{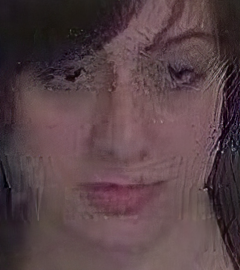}
        \includegraphics[width=\textwidth]{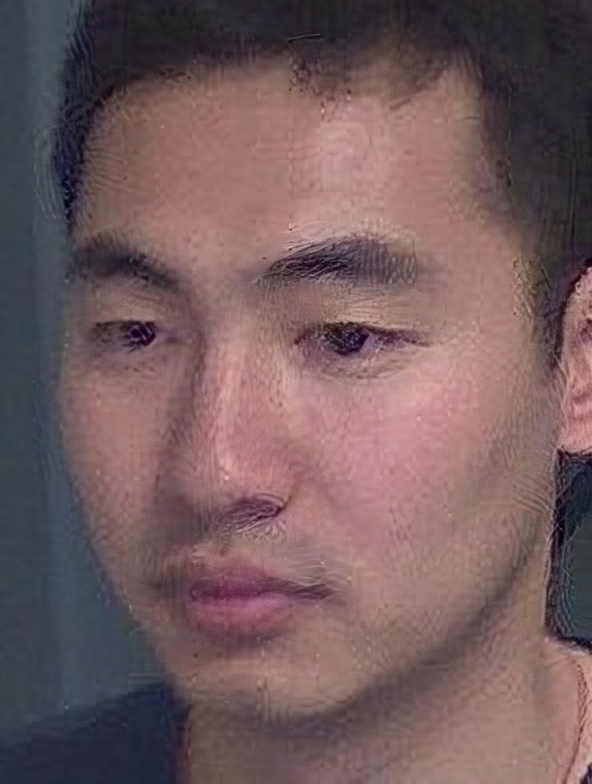}
  \caption*{ESRGAN\cite{esrgan}}
\endminipage\hspace{0.3cm}
\minipage{0.1\textwidth}
    \centering\captionsetup{margin=0.0cm}
        \includegraphics[width=\textwidth]{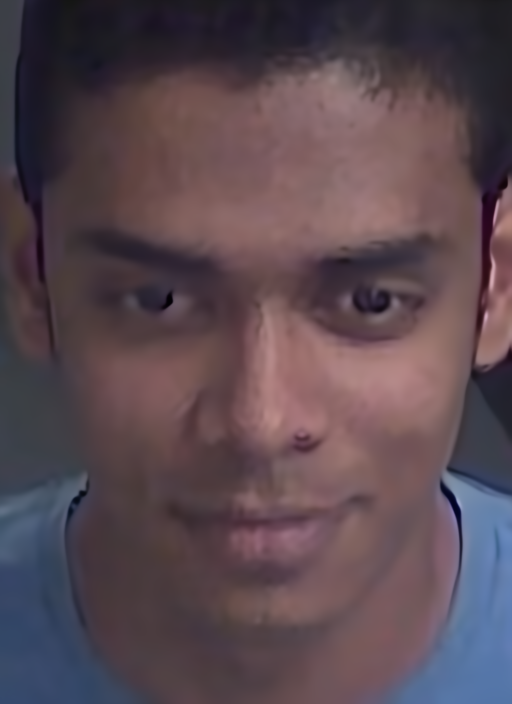}
        \includegraphics[width=\textwidth]{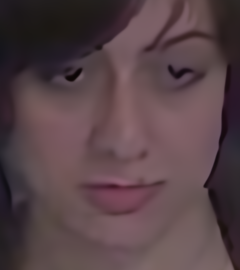}
        \includegraphics[width=\textwidth]{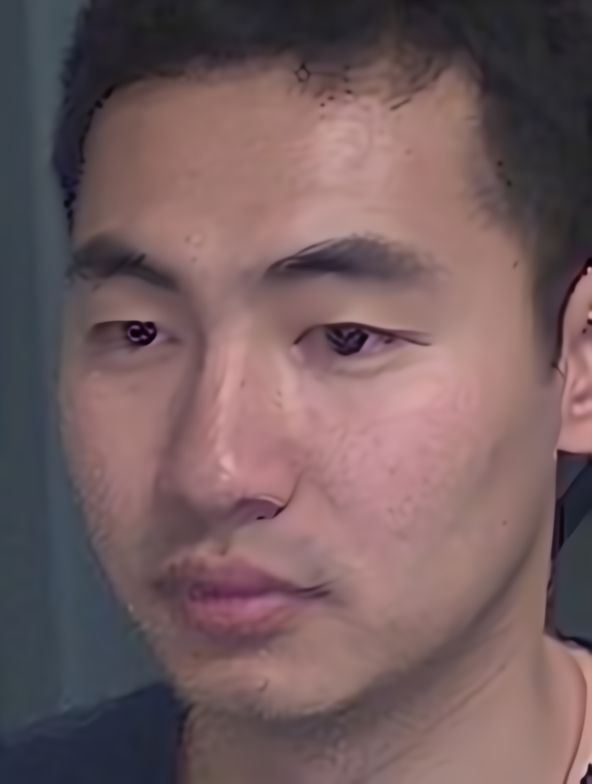}
  \caption*{USRNet\cite{usrnet}}
\endminipage\hspace{0.3cm}
\minipage{0.1\textwidth}
    \centering\captionsetup{margin=0.0cm}
        \includegraphics[width=\textwidth]{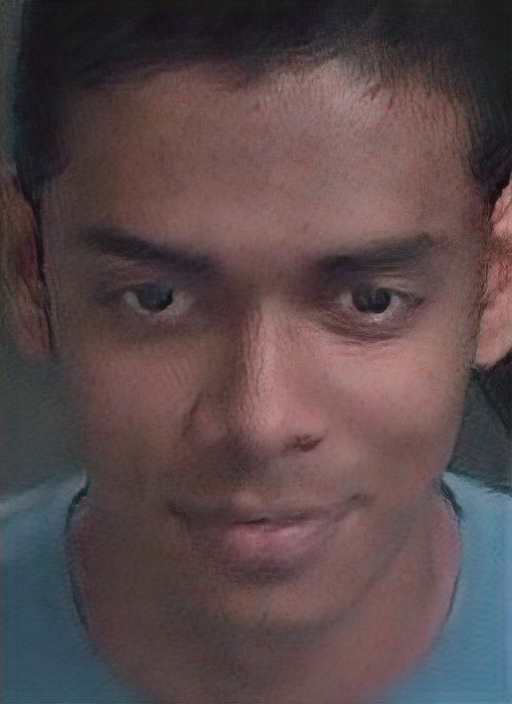}
        \includegraphics[width=\textwidth]{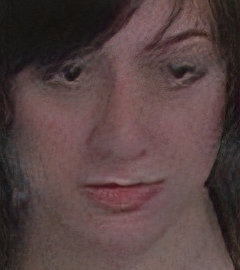}
        \includegraphics[width=\textwidth]{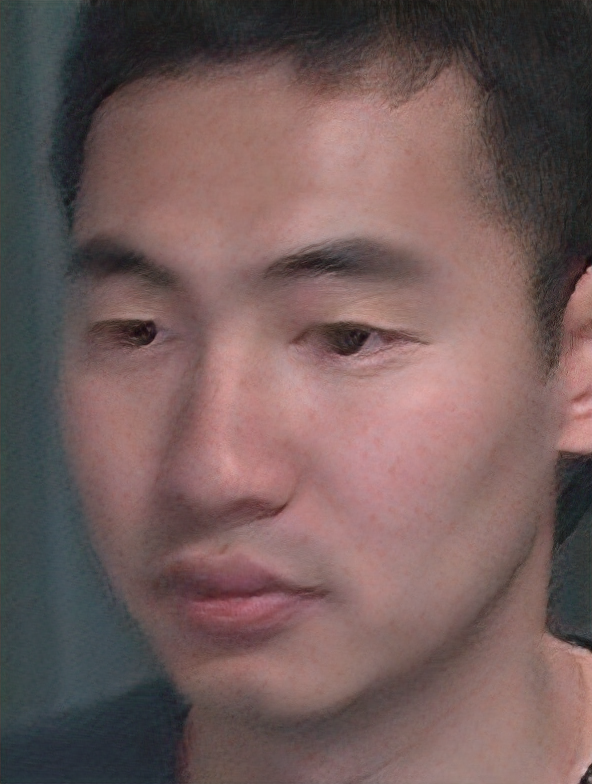}
  \caption*{RealSR\cite{realsr}}
\endminipage\hspace{0.3cm}
\minipage{0.1\textwidth}
    \centering\captionsetup{margin=0.0cm}
        \includegraphics[width=\textwidth]{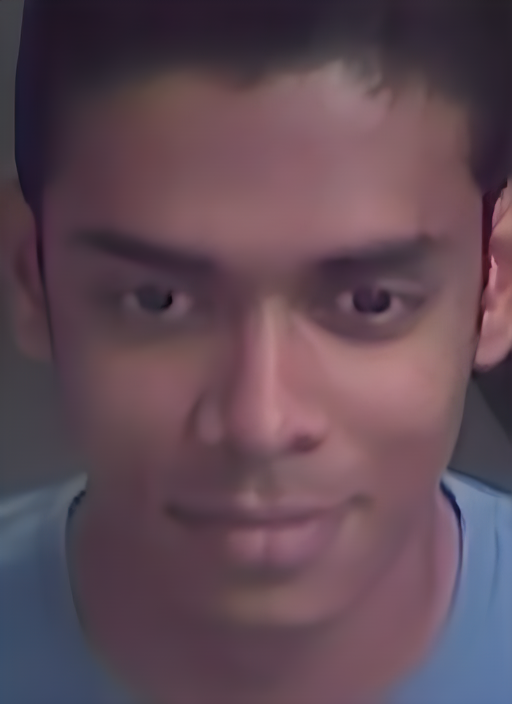}
        \includegraphics[width=\textwidth]{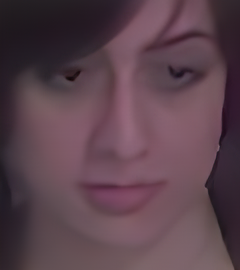}
        \includegraphics[width=\textwidth]{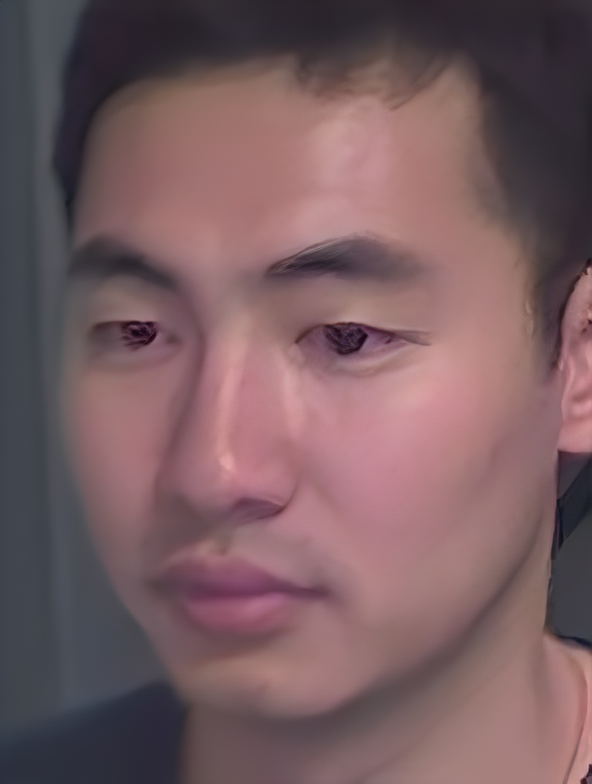}
  \caption*{DPSR\cite{dpsr}}
\endminipage\hspace{0.3cm}
\minipage{0.1\textwidth}
    \centering\captionsetup{margin=0.0cm}
        \includegraphics[width=\textwidth]{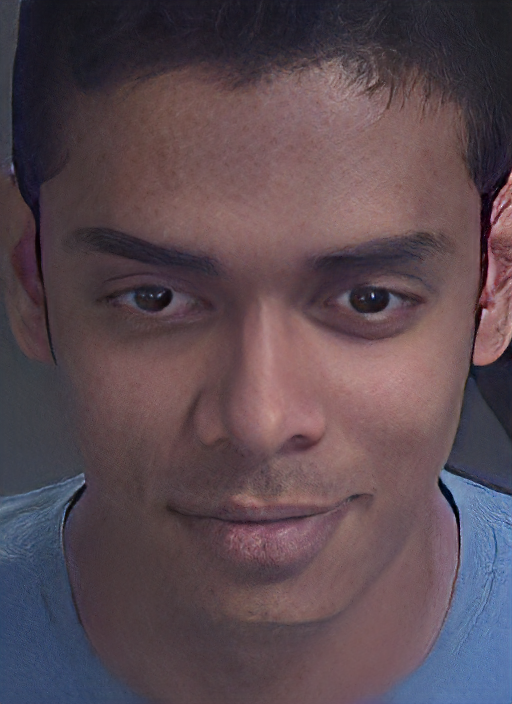}
        \includegraphics[width=\textwidth]{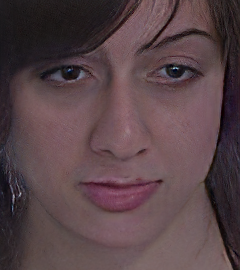}
        \includegraphics[width=\textwidth]{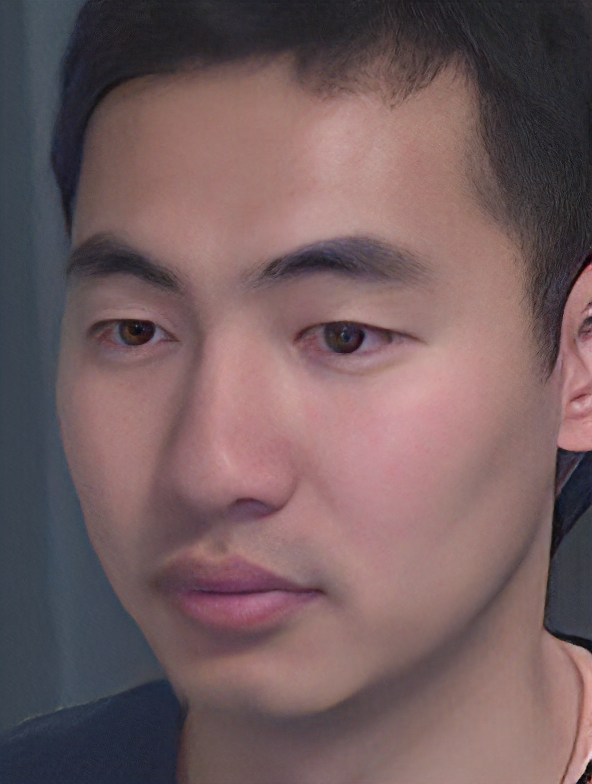}
  \caption*{Ours}
\endminipage
\caption{Comparison with \gls{sota} methods for $\times4$ \gls{sr} of real low-quality face images from the Chokepoint DB \cite{chokepoint}. As visible, our method generates superior reconstructions over the existing methods for different faces.}
\label{fig:comparison1}
\end{figure*}

\begin{figure*}[!h]
\centering
\minipage{0.1\textwidth}
    \centering\captionsetup{margin=0.0cm}
        \includegraphics[width=\textwidth]{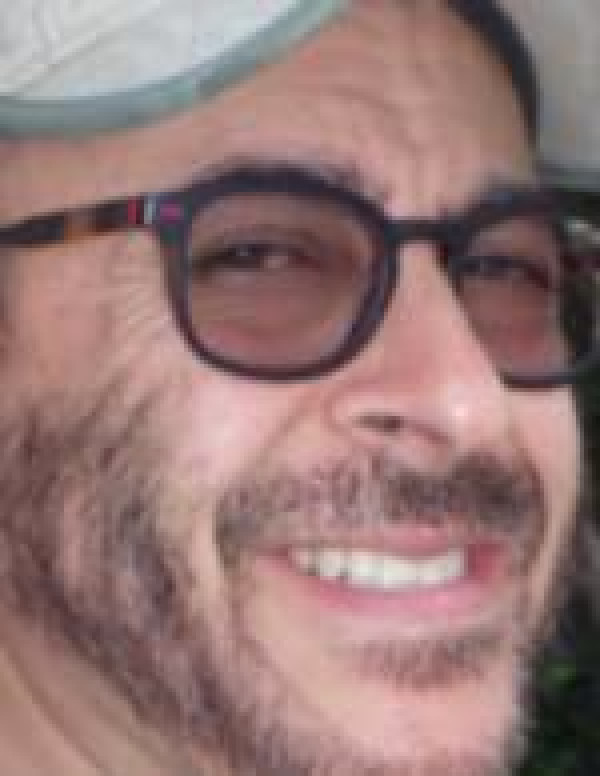}
        \includegraphics[width=\textwidth]{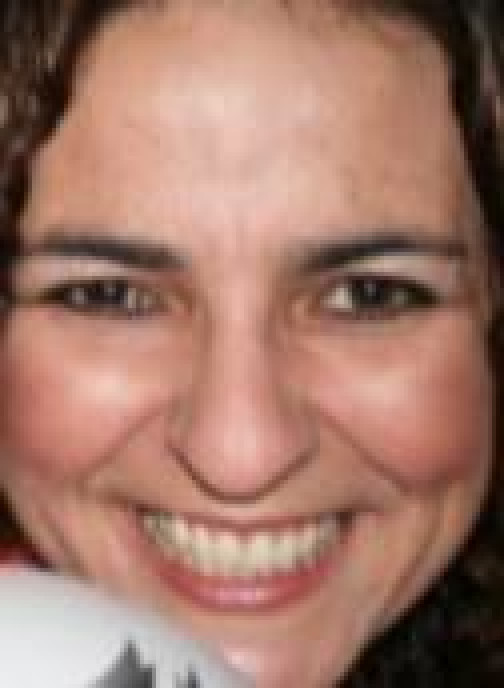}
        \includegraphics[width=\textwidth]{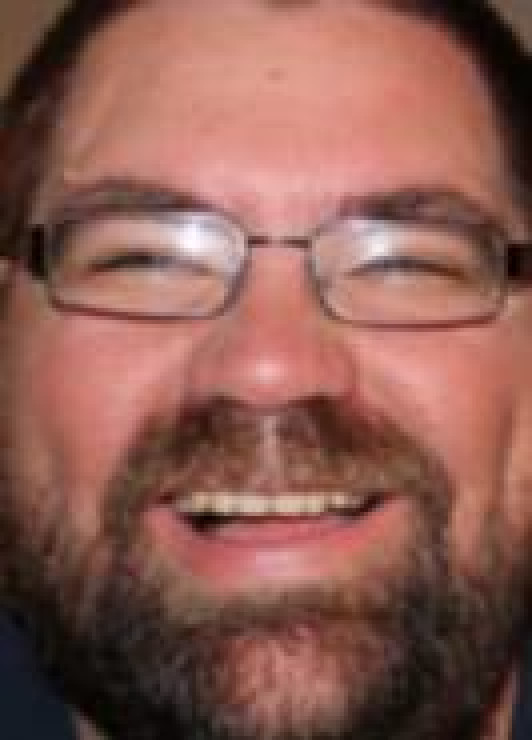}
  \caption*{Original}
\endminipage\hspace{0.3cm}
\minipage{0.1\textwidth}
    \centering\captionsetup{margin=0.0cm}
        \includegraphics[width=\textwidth]{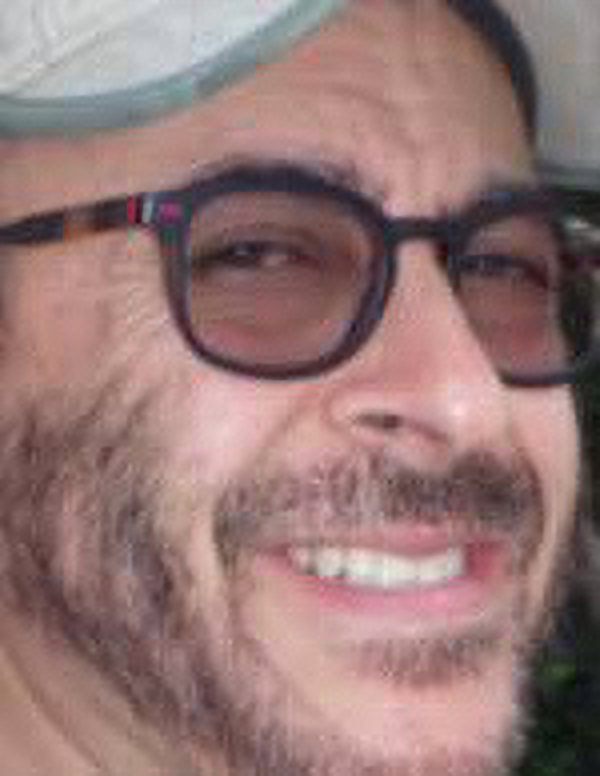}
        \includegraphics[width=\textwidth]{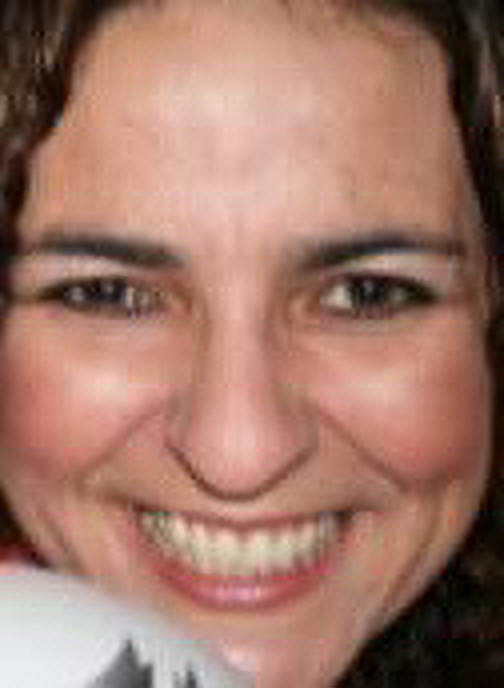}
        \includegraphics[width=\textwidth]{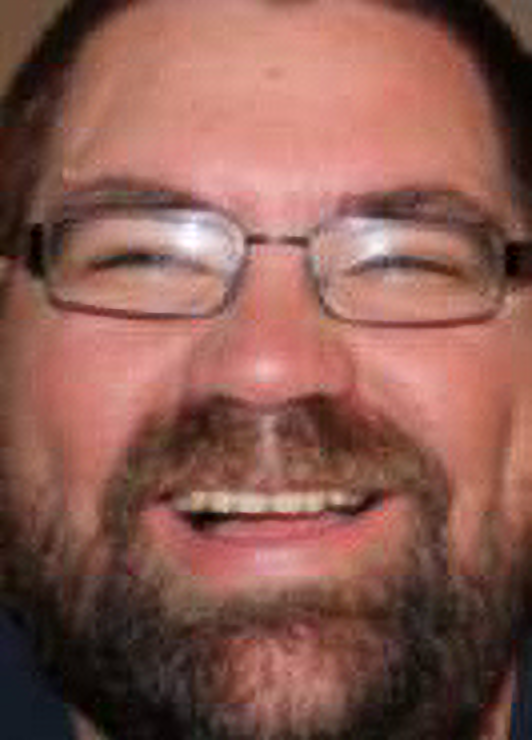}
  \caption*{MZSR\cite{mzsr}}
\endminipage\hspace{0.3cm}
%\minipage{0.11\textwidth}
%        \includegraphics[width=\textwidth]{figs/00716_edsr.png}
%        \includegraphics[width=\textwidth]{figs/00185_edsr.png}
%        \includegraphics[width=\textwidth]{figs/00229_edsr.png}
%        \includegraphics[width=\textwidth]{figs/00023_edsr.png}
%  \caption*{\footnotesize EDSR \cite{edsr}}
%\endminipage\hfill
\minipage{0.1\textwidth}
    \centering\captionsetup{margin=0.0cm}
        \includegraphics[width=\textwidth]{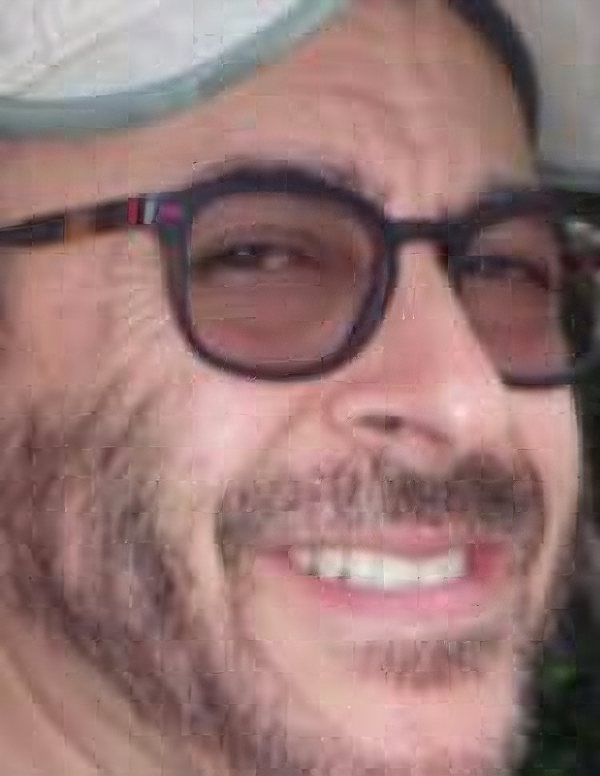}
        \includegraphics[width=\textwidth]{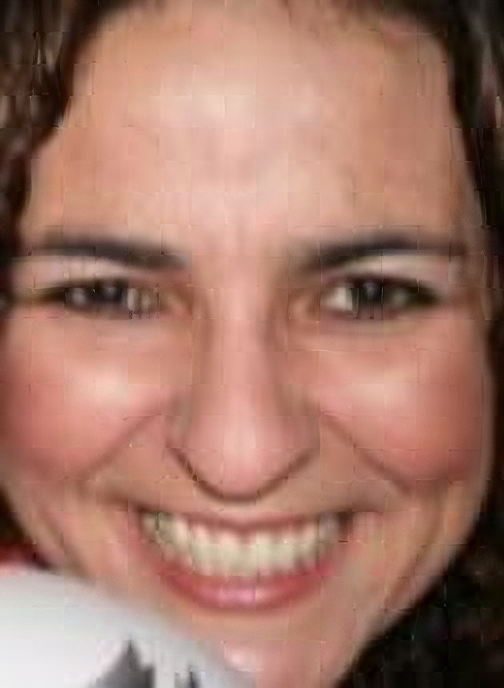}
        \includegraphics[width=\textwidth]{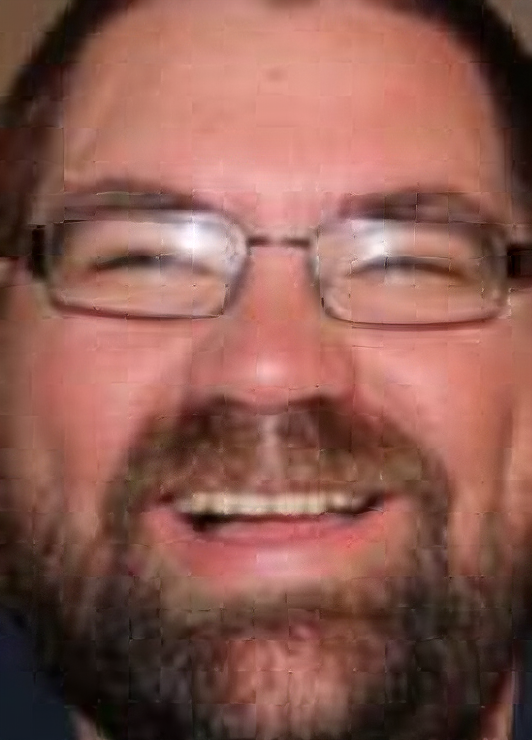}
  \caption*{ESRGAN\cite{esrgan}}
\endminipage\hspace{0.3cm}
\minipage{0.1\textwidth}
    \centering\captionsetup{margin=0.0cm}
        \includegraphics[width=\textwidth]{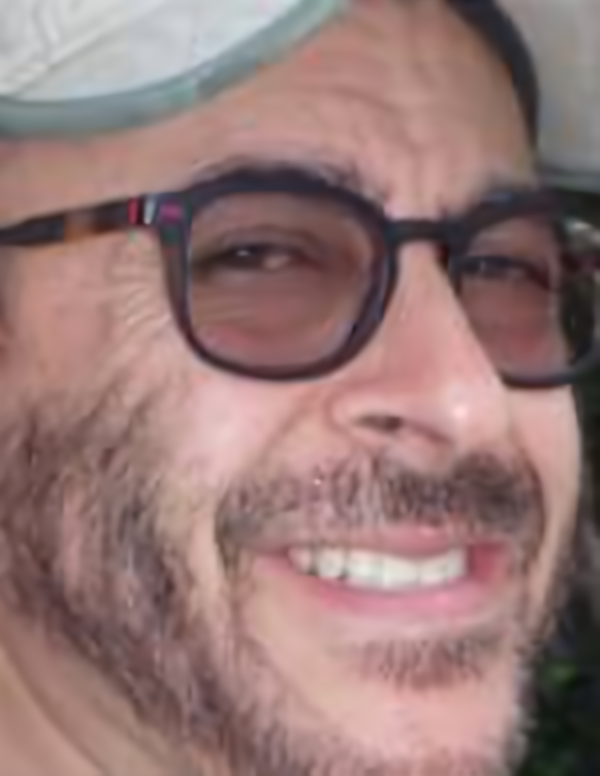}
        \includegraphics[width=\textwidth]{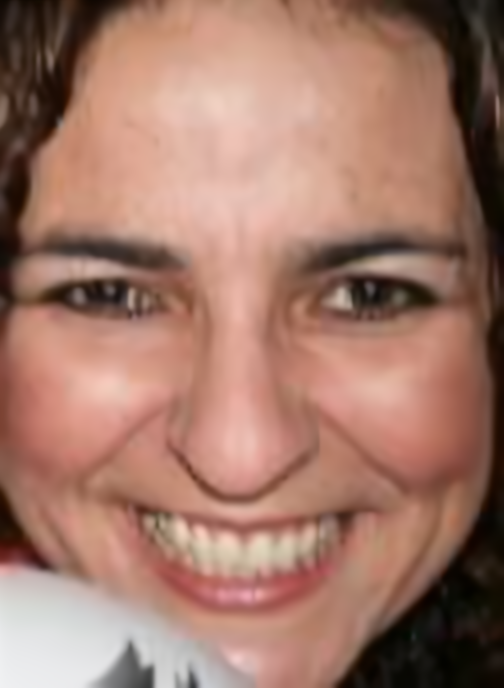}
        \includegraphics[width=\textwidth]{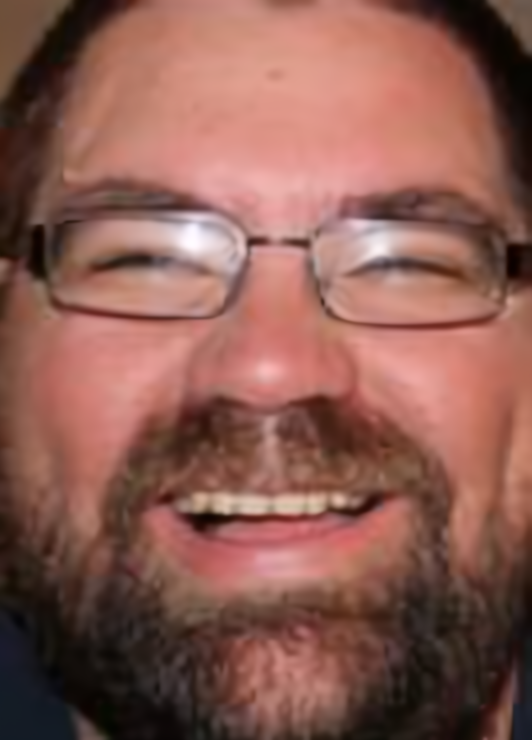}
  \caption*{USRNet\cite{usrnet}}
\endminipage\hspace{0.3cm}
\minipage{0.1\textwidth}
    \centering\captionsetup{margin=0.0cm}
        \includegraphics[width=\textwidth]{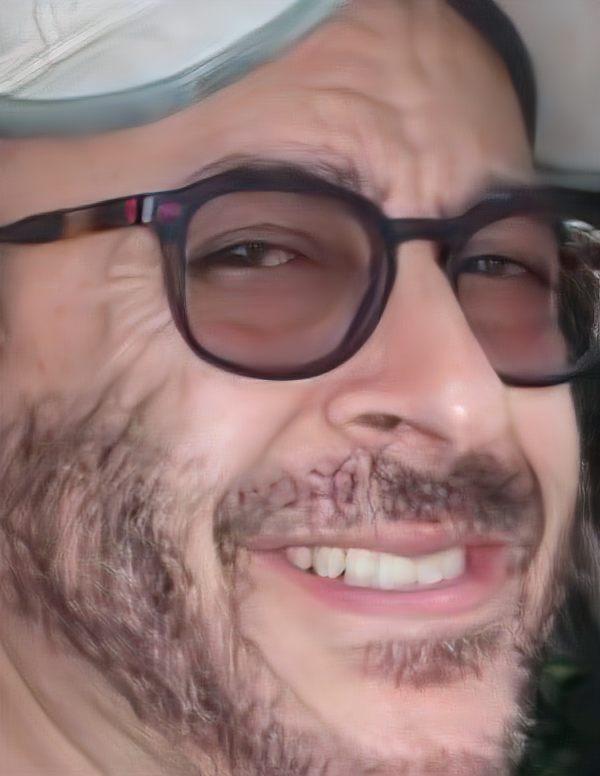}
        \includegraphics[width=\textwidth]{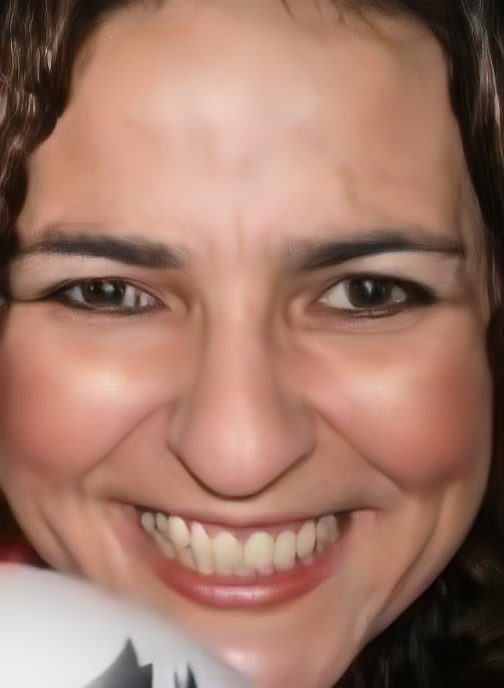}
        \includegraphics[width=\textwidth]{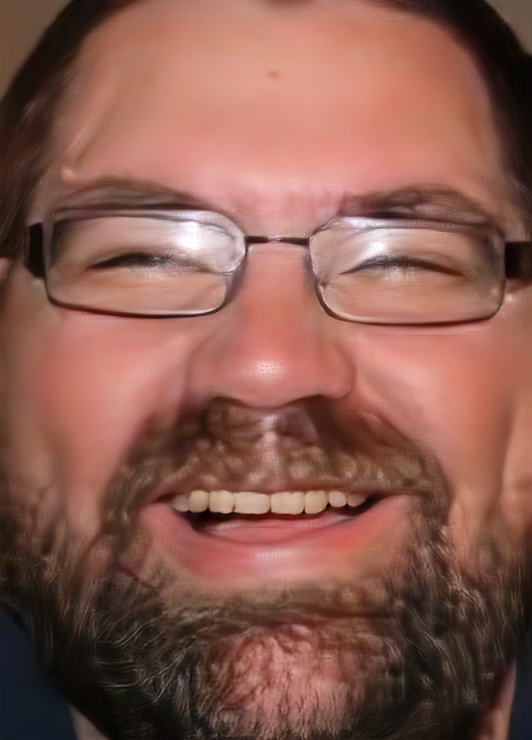}
  \caption*{RealSR\cite{realsr}}
\endminipage\hspace{0.3cm}
\minipage{0.1\textwidth}
    \centering\captionsetup{margin=0.0cm}
        \includegraphics[width=\textwidth]{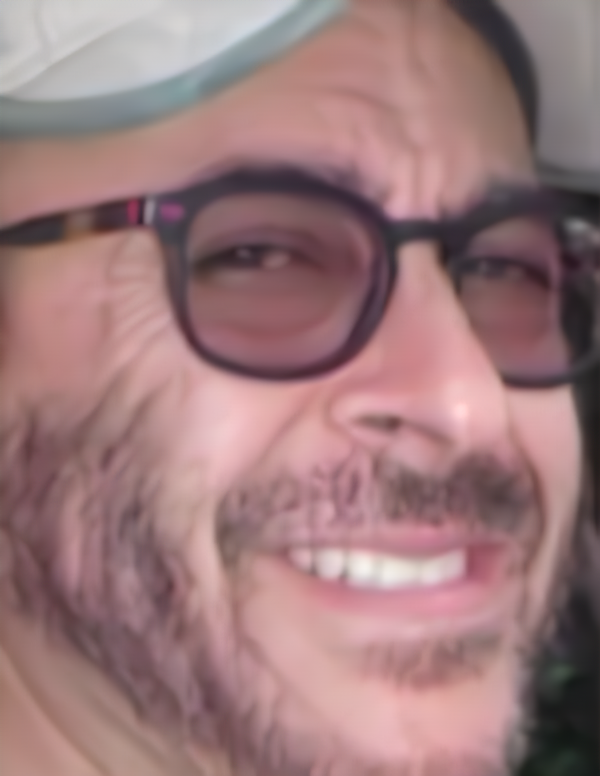}
        \includegraphics[width=\textwidth]{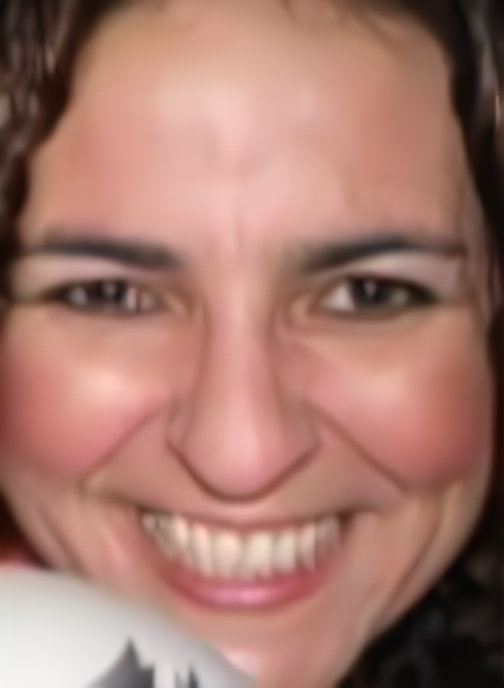}
        \includegraphics[width=\textwidth]{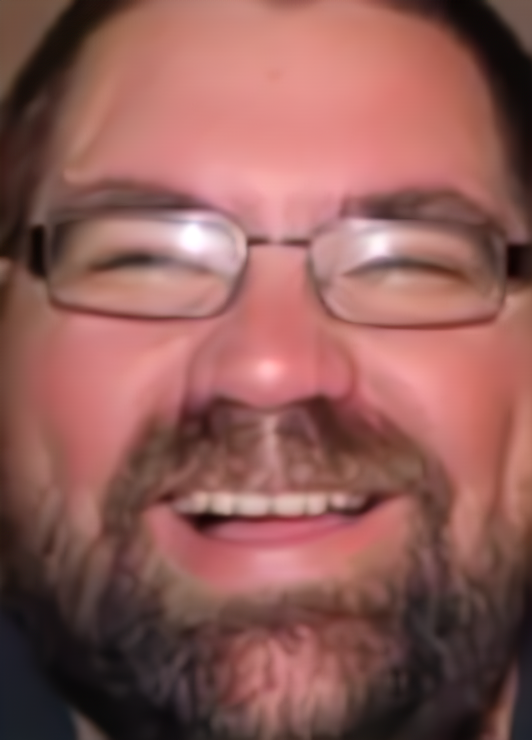}
  \caption*{DPSR\cite{dpsr}}
\endminipage\hspace{0.3cm}
\minipage{0.1\textwidth}
    \centering\captionsetup{margin=0.0cm}
        \includegraphics[width=\textwidth]{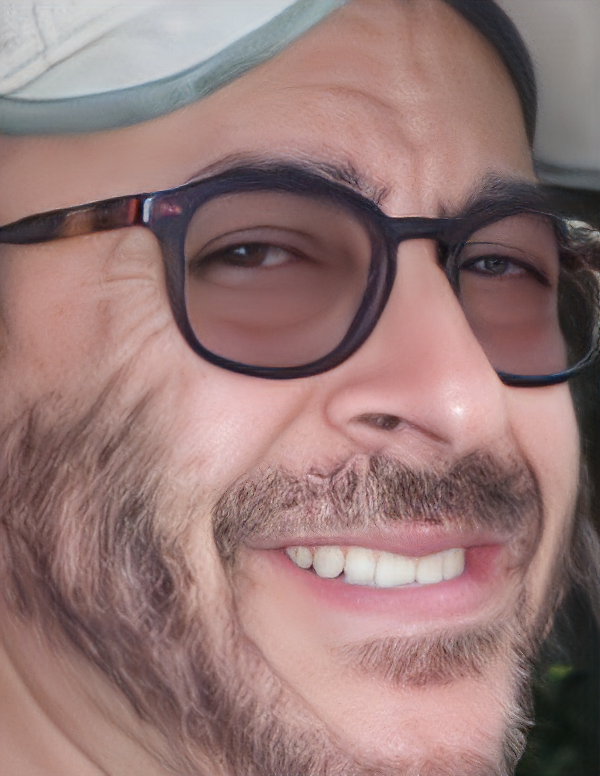}
        \includegraphics[width=\textwidth]{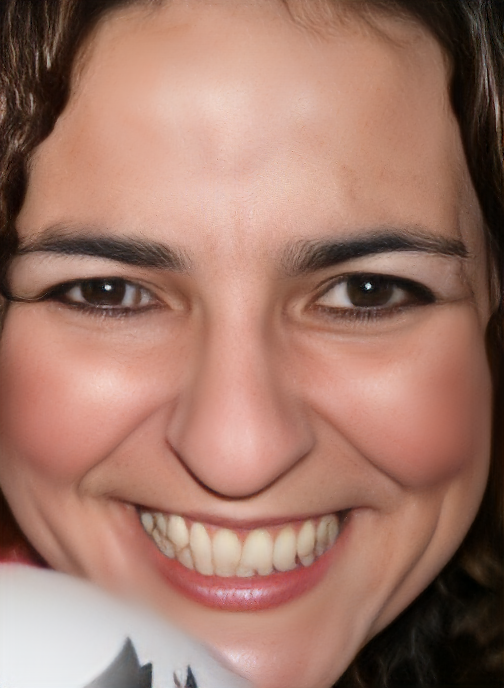}
        \includegraphics[width=\textwidth]{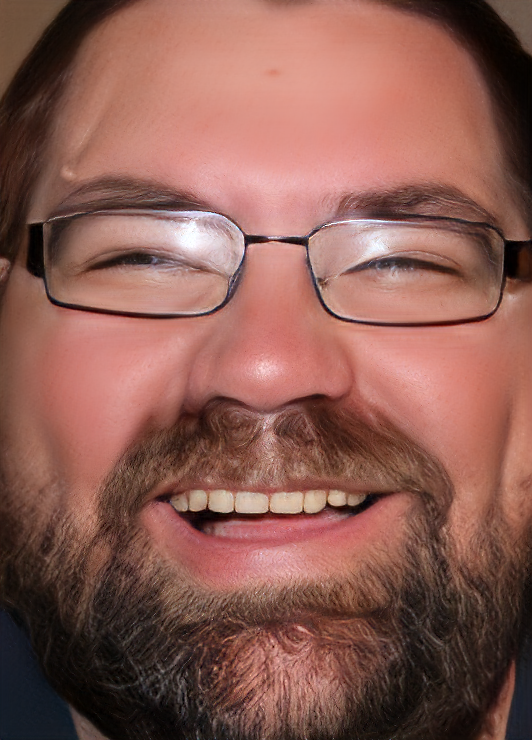}
  \caption*{Ours}
\endminipage\hspace{0.3cm}
\minipage{0.1\textwidth}
    \centering\captionsetup{margin=0.0cm}
        \includegraphics[width=\textwidth]{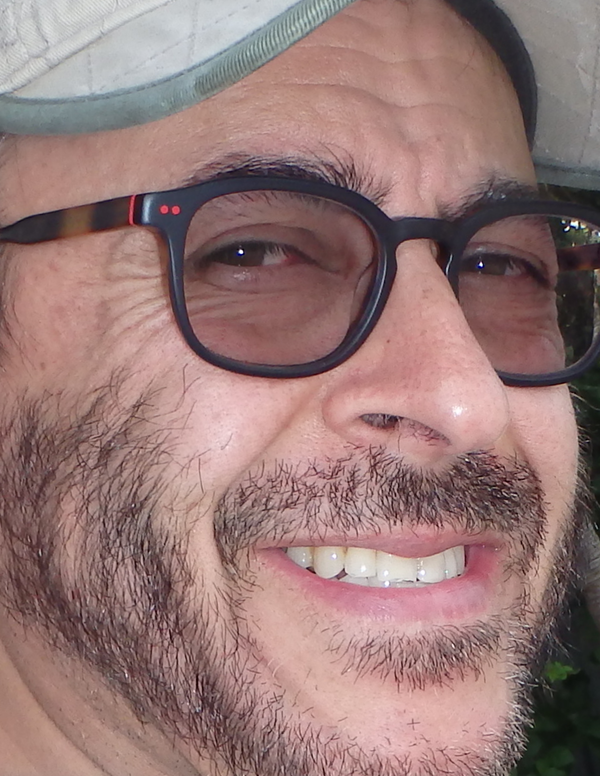}
        \includegraphics[width=\textwidth]{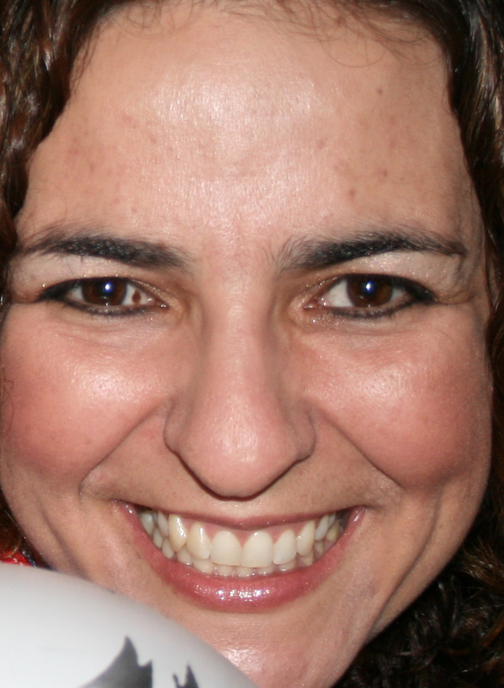}
        \includegraphics[width=\textwidth]{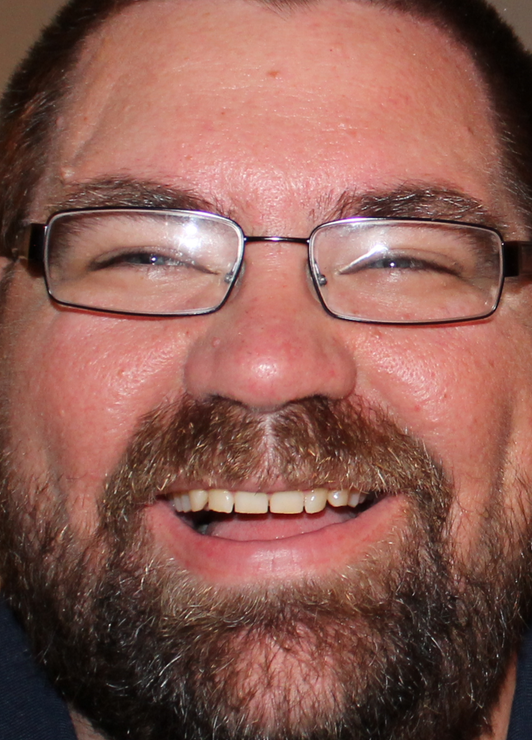}
  \caption*{GT}
\endminipage\
\caption{Comparison with \gls{sota} methods for $\times4$ \gls{sr} of artificially corrupted face images from the \gls{ffhq} \cite{ffhq} testset. As seen, our method hallucinates faces with richer detail and less artifacts compared to the existing methods.}
\label{fig:comparison-synth}
\end{figure*}

\noindent \textbf{Loss Functions}\space\space\space While traditional supervised \gls{sr} models are trained with pixel loss to minimize the \gls{mse} between the reconstructed \gls{hr} image and the \gls{gt} image, we rely on loss functions that maximize the perceptual quality. The original ESRGAN \cite{esrgan} model uses several different loss functions during training. More specifically, the generator uses adversarial loss $\mathcal{L}_{adv}$ \cite{gan} in combination with VGG perceptual loss $\mathcal{L}_{vgg}$ \cite{vggLoss} and pixel loss $\mathcal{L}_{pix}$, while the discriminator use VGG-128 \cite{vggnet} loss $\mathcal{L}_{vgg}$. However, we find that this combination of loss functions is not ideal for high perceptual quality. Following the work of \cite{realsr}, we first exchange the VGG-128 \cite{vggnet} discriminator loss with a PatchGAN discriminator from \cite{cyclegan} to reduce the amount of artifacts in the reconstructed images. Different from the VGG loss, the PatchGAN loss $\mathcal{L}_{patch}$ has a fully convolutional structure, and only penalizes structure differences at the scale of patches, to determine if an image is real or fake. For optimization of the generator, the loss from all patches are averaged and fed back to the generator. Continuing this track, we seek to also replace the VGG-loss in the generator. Inspired by \cite{investigatingLossFunctions}, we find that using the \gls{lpips} perceptual loss $\mathcal{L}_{lpips}$ \cite{lpips} results in less noise and richer textures compared to using VGG-loss for the generator. This is mainly because the VGG network is trained for image classification, while \gls{lpips} is trained to score image patches based on human perceptual similarity judgements. The \gls{lpips} perceptual loss is formulated as:
\begin{equation}
    \mathcal{L}_{lpips} = \sum_{k} \tau^k (\phi^k (I_{gen}) - \phi^k(I_{gt}))
\end{equation}
where $I_{gen}$ is a generated image, $I_{gt}$ is the corresonding \gls{gt} image, $\phi$ is a feature extractor, $\tau$ is a transformation from embeddings to a scalar \gls{lpips} score. The score is computed from $k$ layers and averaged. In our implementation of \gls{lpips} we use the pre-trained AlexNet model provided by the authors. 
In total, our full training loss for the generator is as follows:
\begin{equation}
    \mathcal{L}_{generator} = \lambda_{pix} \cdot \mathcal{L}_{pix} + \lambda_{adv} \cdot \mathcal{L}_{adv} + \lambda_{lpips} \cdot \mathcal{L}_{lpips}
\end{equation}
where $\lambda_{pix}$, $\lambda_{adv}$ and $\lambda_{lpips}$ are scaling parameters.

\subsection{Datasets}
This section describes the datasets used for training and testing. For our experiments on real \gls{lr} face images from surveillance cameras we use the Chokepoint Dataset \cite{chokepoint} as our source domain images $X$. This dataset contains images of 29 different persons captured with three cameras in a real-world surveillance setting. All images have a resolution of $800 \times 600$. We use a face detection algorithm to extract the faces from the images, and randomly split the dataset, to obtain 72,282 images for training and 3,805 images for testing. The average resolution of the cropped faces is $\approx92\times92$. We only use the Chokepoint training images to estimate realistic blur kernels and noise distributions for our degradation framework, and not for direct training of our \gls{sr} model. 

For the target domain of high-quality face images $Y$, we combine 571 face images from the SiblingsDB \cite{siblingsdb}, 8,040 face images from the Radboud Faces Database \cite{raboud-face} and 5,000 randomly selected face images from \gls{ffhq} database \cite{ffhq} for a total of 13,611 images. Both the SiblingsDB and Raboud Face Database contains portrait face images professionally captured in a studio setting with controlled lighting. The face images from the \gls{ffhq} are more diverse in appearance, and ethnicity of the subjects. We augment all images in the target domain by downsampling by 25, 50 and 75\% with bicubic downscaling to obtain a more diverse dataset. We then apply our degradation framework described in Section~\ref{sec:framework} on the images in $Y$ to obtain \gls{lr}/\gls{hr} image pairs for training of our \gls{sr} model. 

For evaluation on artificially corrupted faces images, we use the first 1,000 images from the \gls{ffhq} dataset. To generate \gls{lr}/\gls{gt} images we introduce three kinds of corruptions, namely, downsampling, sensor noise, and compression artifacts. For downsampling, we randomly choose a kernel from our blur kernel pool. For modeling of sensor noise we follow the protocol from \cite{unsupervisedRWSR} and use pixel-wise independent Gaussian noise, with zero mean and a standard deviation of 8 pixels. For compression artifacts, we convert the images to JPEG using a compression strength of 30. 

\begin{table*}[!h]
\begin{center}
\begin{tabular}{|l|c|c|c|c|c|c|c|}
\hline
%\multicolumn{1}{|c}{} &\multicolumn{5}{|c}{Not correlated} & \multicolumn{2}{|c|}{Correlated} \\ \hline
Method & NIQE $\downarrow$ & BRISQUE $\downarrow$ & PIQE $\downarrow$ & NRQM $\uparrow$ & PI $\downarrow$ & NIMA $\uparrow$ & MOR $\downarrow$ \\ \hline
\hline
Bicubic \cite{cubic} & 5.77 & 56.77 & 86.28 & 3.09 & 6.34 & 3.92 & - \\
MZSR \cite{mzsr} & 7.36 & 50.09 & 77.63 & 3.75 & 6.81 & 3.97 & -\\
EDSR \cite{edsr} & 5.43 & 50.63 & 81.97 & 3.82 & 5.81 & 4.08 & - \\
ESRGAN \cite{esrgan} & 3.75 & 19.35 & 19.20 & 7.08 & \textbf{3.34} &4.34 & 4.72\\
USRNet \cite{usrnet} & 6.10 & 59.13 & 87.70 & 3.19 & 6.46 & 4.75 & 3.11\\
RealSR \cite{realsr} & \textbf{3.50} & \textbf{17.20} & \textbf{9.11} & 5.45 & 4.00 & 4.93 & 3.39\\
DPSR \cite{dpsr} & 5.58 & 55.52 & 60.99 & 3.38 & 6.10 & 5.15 & 2.71\\
%Ours$_{VGG}$ & 3.54 & 31.69 & 24.63 & 4.53 & 4.51 &4.98\\ 
Ours & 4.56 & 19.07 & 14.61 & \textbf{7.62} & 3.47 & \textbf{5.92} & \textbf{1.43}\\
\hline
\end{tabular}
\end{center}
\caption{Quantitative results on the Chokepoint testset. $\uparrow$ and $\downarrow$ indicate whether higher or lower values are desired, respectively. Our model scores lower on the traditional \gls{iqa} metrics while being superior on the more recent \gls{nima} metric and \gls{mor} which indicate that the traditional \gls{iqa} metrics are not ideal for evaluation of perceptual quality.}
\label{tab:results}
\end{table*}

\begin{table*}[!h]
\begin{center}
\begin{tabular}{|l|c|c|c|c|c|c|}
\hline
Method & PSNR $\uparrow$ & SSIM $\uparrow$ & MS-SSIM $\uparrow$ & NLPD $\downarrow$ & LPIPS $\downarrow$ & DISTS $\downarrow$ \\
\hline\hline
Bicubic \cite{cubic} & 28.39 & 0.79 & 0.88 & 0.32 & 0.52 & 0.20\\
MZSR \cite{mzsr} & 29.56 & 0.78 & 0.89 & 0.29 & 0.43 & 0.18\\
EDSR \cite{edsr} & 28.27 & 0.78 & 0.88 & 0.33 & 0.50 & 0.19\\
ESRGAN \cite{esrgan} & 28.09 & 0.77 & 0.88 & 0.34 & 0.40 & 0.19\\
USRNet \cite{usrnet} & 28.53 & \textbf{0.80} & 0.89 & 0.32 & 0.53 & 0.21\\
RealSR \cite{realsr} & 29.14 & 0.79 & 0.90 & 0.29 & 0.29 & 0.18 \\
DPSR \cite{dpsr} & 27.45 & 0.79 & 0.88 & 0.33 & 0.51 & 0.25\\
%Ours$_{VGG}$ & 3.54 & 31.69 & 24.63 & 4.53 & 4.51 &4.98\\ 
Ours & \textbf{30.20} & 0.79 & \textbf{0.91} & \textbf{0.28} & \textbf{0.25} & \textbf{0.16}\\
\hline
\end{tabular}
\end{center}
\caption{Quantitative results on the \gls{ffhq} testset. $\uparrow$ and $\downarrow$ indicate whether higher or lower values are desired, respectively.}
\label{tab:results-psnr}
\end{table*}

\subsection{Evaluation Metrics}
\noindent \textbf{Real-World Images}\space\space\space Due to the nature of \gls{rwsr}, no \gls{gt} reference image exists, which makes it impossible to compare the different methods using traditional \gls{sr} \gls{iqa} methods \eg \gls{psnr} and \gls{ssim}. To this end, we follow the non-reference based \gls{iqa} evaluation protocol from the NTIRE2020 \gls{rwsr} challenge \cite{ntire2020}. In particular, we assess the image quality using NIQE \cite{niqe}, BRISQUE \cite{brisque}, PIQE \cite{piqe}, NQRM \cite{ma} and PI \cite{pirm2018}, where PI is a weighted score computed as $\frac{1}{2}((10-NQRM)+NIQE)$. However, these methods are known to correlate poorly with human ratings \cite{ntire2020}. To address this issue, we supplement our evaluation protocol with \gls{mor} and \gls{nima} \cite{nima}, where \gls{nima} is a learned metric based on human opinion scores, which can quantify image quality with high correlation to human perception. We use the pre-trained model for rating of the technical image quality. For the \gls{mor}, we ask the participants to rank overall image quality of the \gls{sr} results. To simplify the ranking, we only include the predictions of the top-5 methods based on \gls{nima} scores. To avoid bias, the order of the methods are randomly shuffled. We average the assigned rank of each method over all images and participants to compute the \gls{mor}.\\

\noindent \textbf{Artificially Corrupted Images}\space\space\space For our experiments on artificially corrupted images we evaluate the performance using three conventional \gls{iqa} methods, \gls{psnr}, \gls{ssim}, and the later \gls{msssim} \cite{msssim}. However, these metrics focus more on signal fidelity rather than perceptual quality \cite{perceptionDistortionTradeoff}. As our method is optimized towards perceptual quality, we also include three of the most recent full-reference metrics targeting perceptual quality, namely \gls{nlpd} \cite{nlpd}, \gls{lpips} \cite{lpips}, and \gls{dists} \cite{dists}. 

\section{Experiments and Results}
\begin{figure*}[!h]
\begin{center}
\minipage{0.2\textwidth}
        \includegraphics[width=\textwidth]{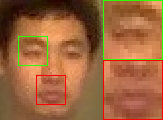}
  \caption*{Original}
\endminipage
\hspace{0.1cm}
\minipage{0.2\textwidth}
        \includegraphics[width=\textwidth]{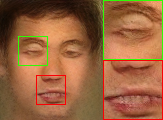}
  \caption*{Baseline}
\endminipage
\hspace{0.1cm}
\minipage{0.2\textwidth}
        \includegraphics[width=\textwidth]{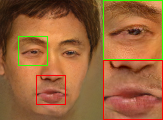}
  \caption*{Compression}
\endminipage
\hspace{0.1cm}
\minipage{0.2\textwidth}
        \includegraphics[width=\textwidth]{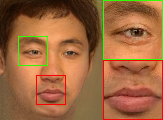}
  \caption*{LPIPS Loss}
\endminipage
\caption{Ablation study of the effect of including compression artifacts in the degradation framework and exchanging the VGG-loss with LPIPS-loss for the generator in the \gls{sr} model, compared to the baseline and the original \gls{lr} image ($56\times64$ pixels)}
\label{fig:ablation}
\end{center}
\vspace{-0.5cm}
\end{figure*}

\noindent \textbf{Implementation Details}\space\space\space
We perform all our experiments with a scaling factor $s=4$. For our \gls{sr} model we jointly train the generator and discriminator for 400K iterations with a batch size of 16. We initialize the weights from the \gls{psnr} optimized RRDB model from \cite{esrgan}. We use \gls{lr} patches of size $32\times32$, and empirically set $\lambda_{pix}$, $\lambda_{adv}$ and $\lambda_{lpips}$ to 0.01, 0.005 and 0.001 respectively. For noise estimation we set $p_i$ to match the \gls{lr} patch size and $q_{j}^{i}$ to 8. Similar to \cite{imageBlindDenoising} we set $\mu$ and $\gamma$ to 0.1 and 0.25 respectively. We empirically set the minimum variance threshold $\phi$ to 0.5. For degradation with compression artifacts we JPEG compress the \gls{lr} training images with strength of 30 during training with a probability of 0.9 for better regularization of the \gls{sr} model.

\subsection{Comparison with State-of-the-Art}
We did not find any other $\times4$ face image specific \gls{rwsr} methods in the literature. Instead, we compare our method to bicubic upscaling, as well as with different groups of \gls{sota} super-resolution methods including two generic SR models (ESRGAN \cite{esrgan}, EDSR \cite{edsr}), one SR method for arbitrary blur kernels (DPSR \cite{dpsr}), three real-world SR models (MZSR \cite{mzsr}, USRNet \cite{usrnet}, and RealSR \cite{realsr}). For a fair comparison, we adjust the competing models for optimal performance. For MZSR \cite{mzsr}, which is an unsupervised method, we enable back-projection with 10 iterations and set a noise level of 0.5. For DPSR \cite{dpsr}, we use the pre-trained DPSRGAN model with settings for real-world images. With USRNet \cite{usrnet} we set the noise value to 15 for best results. The results for the RealSR \cite{realsr}, is based on our re-implementation of the framework as the training code was not available. We adapt the RealSR method to our face data for a fair comparison. For ESRGAN, we use the pre-trained weights provided by the authors to better illustrate the difference from our method.\\

\noindent \textbf{Real-World Images}\space\space\space\label{sec:eval-realworld}  In this experiment we evaluate the \gls{sr} performance on \gls{lr} face images from the Chokepoint testset. Quantitative results can be seen in Table~\ref{tab:results}. Qualitative results for multiple images are shown in Figure~\ref{fig:comparison1} while a close-up view of facial components can be seen in Figure~\ref{fig:closeup}. Our method clearly outperforms the other methods in terms of perceptual quality. However, while the traditional non-reference \gls{iqa} methods (NIQE \cite{niqe}, BRISQUE \cite{brisque}, PIQE \cite{piqe} and NQRM \cite{ma}) fails to capture this, scores from the more recent NIMA \cite{nima} method correlates well with human perception, which is also backed by our \gls{mor} rankings. This shows that the traditional \gls{iqa} metrics are not ideal for judgement of the perceptual quality.\\

\noindent \textbf{Artificially Corrupted Images}\space\space\space This experiment evaluate the \gls{sr} performance on artificially corrupted images from the \gls{ffhq} testset. We show quantitative results of all methods in Table~\ref{tab:results-psnr}. Qualitative results for multiple images are shown in Figure~\ref{fig:comparison-synth}. Our method produces sharp and detailed images with few artifacts which closely resembles the \gls{gt} images, which is also reflected in the quantitative results. Most noteworthy are the \gls{dists} results, which are very correlated with human perception of image quality. The results show that the reconstructed images produced by our method is superior in comparison to the other methods.

\subsection{Ablation Study}
We evaluate the effect of our proposed method for realistic image degradation and our improved ESRGAN based \gls{sr} model in the same setting as described in Section~\ref{sec:eval-realworld}. A qualitative comparison can be seen in Figure~\ref{fig:ablation}. \\

\noindent \textbf{Baseline}\space\space\space Here, we use kernel estimation and noise injection to generate training data for the ESRGAN with patch discriminator, similar to \cite{realsr}. This \gls{sr} model is fine-tuned to our face image dataset, and serves as our baseline. The resulting \gls{hr} images contain unpleasing noise and lack detail.\\ 

\noindent \textbf{Compression Artifacts}\space\space\space In this setting, we add JPEG compression artifacts to the \gls{lr} images during training of the baseline model. This results in more noise-free reconstructions compared to the baseline.\\

\noindent \textbf{LPIPS loss}\space\space\space Here, we use the LPIPS loss function for the generator instead of VGG-loss combined with the addition of compression artifacts. When the baseline model is re-trained under these settings the resulting reconstructions becomes sharper with better texture and details.

\subsection{Failure Cases}

\begin{figure}[!h]
\minipage{0.11\textwidth}
        \includegraphics[width=\textwidth]{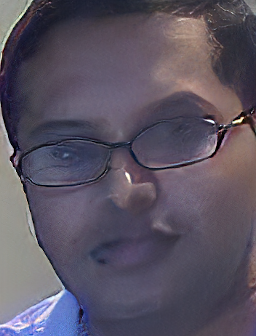}
  \caption*{(a)}
\endminipage\hfill
\minipage{0.11\textwidth}
        \includegraphics[width=\textwidth]{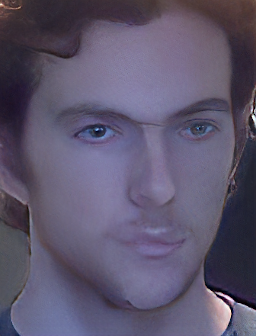}
  \caption*{(b)}
\endminipage\hfill
\minipage{0.11\textwidth}
        \includegraphics[width=\textwidth]{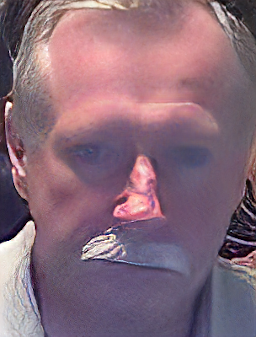}
  \caption*{(c)}
\endminipage\hfill
\minipage{0.11\textwidth}
        \includegraphics[width=\textwidth]{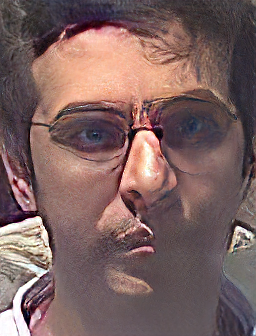}
  \caption*{(d)}
\endminipage
\caption{Examples of failure cases. Figure (a) and (b) illustrate cases where only parts of the image is super-resolved. Figure (c) shows a case where almost no high-frequency details are restored. Figure (d) shows a case where unrealistic facial features are introduced.}
\label{fig:failureCases}
\end{figure}

While our method produces reconstructed faces of better visual quality than the compared \gls{sota} methods, it does not solve the problem \gls{rwsr} of face images. Figure~\ref{fig:failureCases} shows several failure cases of our method. These occur when the input image is severely corrupted \eg by motion blur or harsh lighting, or when out-of-focus. In these cases, our method might only super-resolve some parts of the face, \eg a single eye, or even hallucinate unrealistic facial features.

\section{Conclusion}
In this paper, we have presented a novel framework for \gls{rwsr}, which we have evaluated on low-quality face images from surveillance cameras, and artificially corrupted face images. Our method shows \gls{sota} performance in both cases, which is achieved by using \gls{lpips}-loss and making the \gls{sr} model robust against the most common degradation types present in real \gls{lr} images. Moreover, our model is the first to perform \gls{sr} on real \gls{lr} face images of arbitrary sizes, which makes it useful for practical applications. In the future, even better reconstructions could possibly be obtained by including more image degradation types in the framework \eg chromatic aberration.  

\section{Acknowledgments}
This work was supported by Danmarks Frie Forskningsfond under grant number 8022-00360B, and the Milestone Research Programme at Aalborg University (MRPA).

{\small
\bibliographystyle{ieee_fullname}
\bibliography{bibtex}
}

\end{document}